\documentclass{article}

\usepackage[final]{neurips_2025}

\usepackage[utf8]{inputenc} 
\usepackage[T1]{fontenc}    
\usepackage{hyperref}       
\usepackage{url}            
\usepackage{booktabs}       
\usepackage{amsfonts}       
\usepackage{nicefrac}       
\usepackage{microtype}      
\usepackage{xcolor}         

\usepackage{multirow}
\usepackage{multicol}
\usepackage{graphicx}
\usepackage{subfigure}
\usepackage{wrapfig}
\usepackage{algorithm}
\usepackage{algpseudocode}
\usepackage{authblk}
\usepackage{subcaption}

\usepackage{amssymb, amsmath}
\usepackage{dsfont}
\usepackage{cleveref}

\newtheorem{assumption}{Assumption}

\newtheorem{lemma}{Lemma}
\newtheorem{proof}{Proof}
\makeatletter
\newcommand{\model}{%
  InfMasking
  \futurelet\next@token\check@next@token
}
\newcommand{\check@next@token}{%
  \ifcat a\next@token
    \ 
  \else
    \ifcat 0\next@token
      \ 
    \else
      \ifcat \noexpand~\next@token
      \else
      \fi
    \fi
  \fi
}
\makeatother

\title{InfMasking: Unleashing Synergistic Information \\ by Contrastive Multimodal Interactions}

\author{
\textbf{Liangjian Wen$^{1,2,8\ast}$, Qun Dai$^{1\ast}$,  Jianzhuang Liu$^{3}$, Jiangtao Zheng$^{1}$, Yong Dai$^{4\ddagger}$,Dongkai Wang$^{1\dagger}$, Zhao Kang$^{5}$, Jun Wang$^{1}$, Zenglin Xu$^{6,7}$, Jiang Duan$^{1\dagger}$} \\

$^1$\textmd{School of Computing and Artificial Intelligence, Southwestern University of Finance and Economics}\\
$^2$Engineering Research Center of Intelligent Finance, Ministry of Education,Southwestern University of Finance and Economics\\
$^3$Shenzhen Institutes of Advanced Technology, Chinese Academy of Sciences \\
$^4$X-Humanoid \\
$^5$University of Electronic Science and Technology of China\\
$^6$Shanghai Academy of AI for Science \\
$^7$Artificial Intelligence Innovation and Incubation Institute, Fudan University \\
$^8$Artificial Intelligence and Digital Finance Key Laboratory of Sichuan Province\\
\texttt{\{wlj6816,daiqun1124\}@gmail.com} \\
}
\begin{document}
\maketitle
\renewcommand{\thefootnote}{\fnsymbol{footnote}}
    \footnotetext[1]{Equal contribution.}
    \footnotetext[2]{Corresponding author.}
    \footnotetext[3]{Project leader.}

\begin{abstract}
In multimodal representation learning, synergistic interactions between modalities not only provide complementary information but also create unique outcomes through specific interaction patterns that no single modality could achieve alone.
Existing methods may struggle to effectively capture the full spectrum of synergistic information, leading to suboptimal performance in tasks where such interactions are critical. This is particularly problematic because synergistic information constitutes the fundamental value proposition of multimodal representation. To address this challenge, we introduce {\model}, a contrastive synergistic information extraction method designed to enhance synergistic information through an \textbf{Inf}inite \textbf{Masking} strategy. InfMasking stochastically occludes most features from each modality during fusion, preserving only partial information to create representations with varied synergistic patterns. Unmasked fused representations are then aligned with masked ones through mutual information maximization to encode comprehensive synergistic information. This infinite masking strategy enables capturing richer interactions by exposing the model to diverse partial modality combinations during training. As computing mutual information estimates with infinite masking is computationally prohibitive, we derive an InfMasking loss to approximate this calculation.
Through controlled experiments, we demonstrate that \model effectively enhances synergistic information between modalities. In evaluations on large-scale real-world datasets, \model achieves state-of-the-art performance across seven benchmarks.
Code is released at \url{https://github.com/brightest66/InfMasking}.
\end{abstract}

\section{Introduction}
Multimodal contrastive learning has revolutionized representation learning by enabling the integration of diverse data modalities, such as text, images, and audio, into a unified latent space. This paradigm leverages contrastive loss~\cite{oord2018representation, chen2020simple} to align features from different modalities as pioneered by foundational works like CLIP~\cite{radford2021learning} and ALIGN~\cite{jia2021scaling} in vision-language tasks. These models demonstrate the power of aligning multimodal features to capture shared patterns across data sources, enabling versatile downstream applications. 
However, current approaches often rely on the restrictive \textit{multiview redundancy assumption}~\cite{DBLP:conf/colt/SridharanK08, DBLP:conf/iclr/Tsai0SM21, 10477543}, which posits that a modality is (approximately) sufficient for
the prediction of downstream tasks and contains the same
task-relevant information. This assumption derives from multi-view learning and is limited in real-world multimodal settings because many multimodal tasks involve minimal shared information.

The shortcomings of this redundancy-centric perspective become increasingly apparent when we examine the multifaceted nature of multimodal interactions. As illustrated in  ~\cite{bertschinger2014quantifying,dufumier2024align}, these interactions can be classified into three fundamental categories: redundancy, uniqueness, and synergy. Redundancy refers to scenarios where a modality can independently perform the task due to overlapping, shared information. Uniqueness describes cases where only one modality possesses all the requisite information for task completion. Synergy, arguably the most significant yet elusive of the three, occurs when modalities provide complementary information that must be combined to achieve the desired outcome. These interaction types are not static; their predominance varies depending on the specific task, adding a layer of complexity to multimodal learning. For instance, a task might rely heavily on redundant information in one context, while another demands the synergistic integration of modalities to succeed. 
A compelling example of this is hateful meme detection~\cite{kiela2020hateful}, where synergy emerges when seemingly neutral modalities (such as an innocuous image and benign text) combine to create harmful content that neither conveys on its own. This highlights the critical importance of synergistic integration, as models must fuse information from different modalities cues to uncover implicit biases or offensive implications that are only apparent in their joint context, enabling more effective identification and mitigation of such content in real-world applications.
Consequently, task-agnostic multimodal representations must necessarily encompass the full spectrum of multimodal interactions that transcend mere informational redundancy.

Recently, FactorCL~\cite{liang2023factorized} explicitly decomposes shared and unique representations, enabling the estimation of redundancy and unique interactions beyond multi-view redundancy. 
However, its factorized approximation of multimodal interactions is prone to cumulative errors and fails to capture synergistic information effectively.
In contrast, ~\cite{dufumier2024align} integrates features from all modalities to derive a common representation and subsequently maximizes the mutual information between this representation and its augmented variants, as well as between the common representation and its corresponding unimodal features. 
Although this approach facilitates the capture of redundant, unique, and synergistic information across modalities, it primarily emphasizes enhanced redundant and unique interactions. Synergistic information is captured by maximizing the mutual information between the common representation and its augmented variants. 
Such handling may prove inadequate for tasks that heavily rely on complex inter-modal synergy. The complexity of synergistic interactions lies not merely in modalities providing complementary information but in how these modalities combine through specific interaction patterns to produce outcomes unattainable by any single modality alone.
Such interactions may involve nonlinear dependencies or context-dependent dynamics.
Hence, effectively capturing the full spectrum of synergistic information remains a significant challenge.

To address this challenge, we introduce a contrastive synergistic information extraction method via infinite masking (denoted as InfMasking). Specifically, InfMasking stochastically occludes a substantial proportion of features from each modality during the fusion process.
This masking preserves only partial information, creating fused representations with varied synergistic patterns.
Subsequently, unmasked fused representations are aligned with these masked ones via mutual information maximization to encode comprehensive synergistic information. The infinite masking strategy enables InfMasking to capture richer synergistic interactions by exposing the model to diverse combinations of partial modality information during training. However, the computation of mutual information estimates with infinite masking is computationally prohibitive. To address this issue, we derive an InfMasking loss to approximate the calculation of this loss function. Empirically, InfMasking effectively captures synergistic information between modalities in controlled environments with known interaction types. When tested on real-world datasets across diverse domains (healthcare, robotics) and data types (image, text, audio), InfMasking achieves state-of-the-art results on seven multimodal tasks involving two or three modalities.

\section{Preliminary: Contrastive Multimodal  Interactions }
Consider \(X_1, X_2, \ldots, X_n\) as random variables, each representing a distinct modal data (e.g., image, text, audio, or tabular data), alongside a downstream task \(Y\). 
The objective is to derive  an effective multimodal latent variable \(Z_{\theta} = f_{\theta}(X)\), where \(X = (X_1, \ldots, X_n)\) and $\theta$ detnotes the parameter of multimodal encoder.
Multimodal interactions refer to the process of extracting and integrating information from multiple data sources, such as text, image, audio, or tabular data, to form a cohesive representation for downstream tasks. These interactions can be categorized into three types: redundancy (R), where information is shared across modalities; uniqueness (U), where information is specific to a single modality; and synergy (S), where complementary information emerges only when modalities are combined.

To understand multimodal interactions, we can leverage Partial Information Decomposition (PID)~\cite{williams2010nonnegative,bertschinger2014quantifying}, a theoretical framework that dissects the mutual information between multiple variables. For analytical tractability, we focus on the case of $n = 2$. Consider two modalities \(X_1\) and \(X_2\) and a task \(Y\).
The mutual information \(I(X_1, X_2; Y)\) quantifies the total task-relevant information provided by both modalities jointly. According to PID, this can be decomposed as:
$I(X_1, X_2; Y) = R + S + U_1 + U_2$, where \(R\) represents redundant information, common to both \(X_1\) and \(X_2\), \(S\) represents synergistic information, emerging only from the combination of \(X_1\) and \(X_2\), and \(U_1\) and \(U_2\) represent unique information specific to \(X_1\) and \(X_2\), respectively.
This decomposition is supported by consistency equations derived from the chain rule of mutual information:\[I(X_1; Y) = R + U_1, \quad I(X_2; Y) = R + U_2, \quad I(X_1;X_2; Y) = R - S.\]
Effectively capturing these interactions is fundamental to multimodal learning, as they collectively enhance a model's ability to generalize across diverse tasks.
Therefore, an effective multimodal representation $Z_{\theta}$ must capture information relevant to a downstream task $Y$, preserving the mutual information such that $I(Z;Y) = I(X;Y)$.  In self-supervised learning, however,  $Y$ remains unspecified during the representation learning phase, presenting a unique challenge.
To address this, \cite{dufumier2024align} extends the contrastive learning principle of multiview redundancy to multimodal contexts:
\begin{assumption}[\textit{Minimal label-preserving multimodal augmentations}]
\vspace{-.6em}
\label{assumption: label-preserving}
   A set \( \mathcal{T}^* \) of multimodal transformations exists, such that for any \( t \in \mathcal{T}^* \) and \( X' = t(X) \), the mutual information \( I(X;X') = I(X;Y) \) holds, preserving the information with label \( Y \).
\vspace{-.6em}
\end{assumption}
Assumption~\ref{assumption: label-preserving} is justified within the framework of multimodal representation learning. It enables a broader range of augmentations, as it is not limited to the set $\mathcal{T}_c^{\star}=\left\{t(X)=\left(t_1\left(X_1\right), t_2\left(X_2\right)\right), \ldots, t_n(X_n) \right\}$. We define $Z_{\theta}'= f_{\theta}(X')$, where $X^{\prime}=t(X)$ with $t \in \mathcal{T}$ representing multimodal augmentation. Considering the data processing inequalities for the Markov chains $X \rightarrow X^{\prime} \rightarrow Z_\theta^{\prime}$ and $Z_\theta^{\prime} \rightarrow X \rightarrow Z_\theta$, we can establish the following mutual information bounds:$I\left(Z_\theta ; Z_\theta^{\prime}\right) \leq I\left(X; Z_\theta^{\prime}\right) \leq I\left(X; X^{\prime}\right) $.
According to these inequalities and Assumption~\ref{assumption: label-preserving}, we can prove the following lemmas:
\begin{lemma}\label{lemma: Ixxp}
\vspace{-.6em}
    When optimizing the function $f_\theta$ to maximize mutual information $I\left(Z_\theta; Z_\theta^{\prime}\right)$, and under the assumption that the network $f_\theta$ possesses sufficient expressivity, we observe that in the optimal parameter configuration:   $I(Z_{\theta^{\star}}, Z'_{\theta^{\star}})=I(X, X')=I(X,Y)$.
\vspace{-.6em}
\end{lemma}
For $Z_{\theta}$ to serve as an effective representation of $X$, it must adequately preserve and encode all task-relevant information inherent in $X$. We note that $I(X; Y)=I(X_1, X_2; Y) = R + S + U_1 + U_2$.
Based on Lemma \ref{lemma: Ixxp}, we can learn common multimodal representations $Z_{\theta}$ and quantify all multimodal interactions beyond redundancy by maximizing the mutual information $I\left(Z_\theta; Z_\theta^{\prime}\right)$.
\begin{lemma}\label{cor: Ix1z}
\vspace{-.6em}
Suppose \( f_{{\theta}^*} \) is optimal, meaning it maximizes $I(Z_{\theta^{\star}}, Z'_{\theta^{\star}})$ . Then, the equality \( I(Z_{\theta^{\star}}, Y) = I(X', Y) \) holds. For the special case where \( T = \{t_i\} \) such that \( X' = t_i(X) = X_i \) and \( Z_{\theta^{\star}}' = f_{\theta^{\star}}(X) = Z_i \) for \( i \in \{1, 2\} \), the following holds:$  I(Z_i; Y) = I(X_i;Y) = R+U_i$.
\vspace{-.6em}
\end{lemma}
For unimodal representations $Z_{i}$ where $i \in \{1, 2\}$ to effectively represent $X_i$, each representation must encode the information $I(X_i; Y) = R + U_i$ contained in the corresponding modality. 
According to Lemma \ref{cor: Ix1z}, we can learn optimal unimodal representations $Z_{i}$ and quantify both redundant and unique multimodal interactions by maximizing the mutual information $I\left(Z_i; Z_\theta^{\prime}\right)$ and $I\left(Z_i; Z_\theta\right)$.

\section{Unleashing Synergistic Information through Infinite
Masking}

\begin{figure}
    \centering
    \includegraphics[width=1\linewidth]{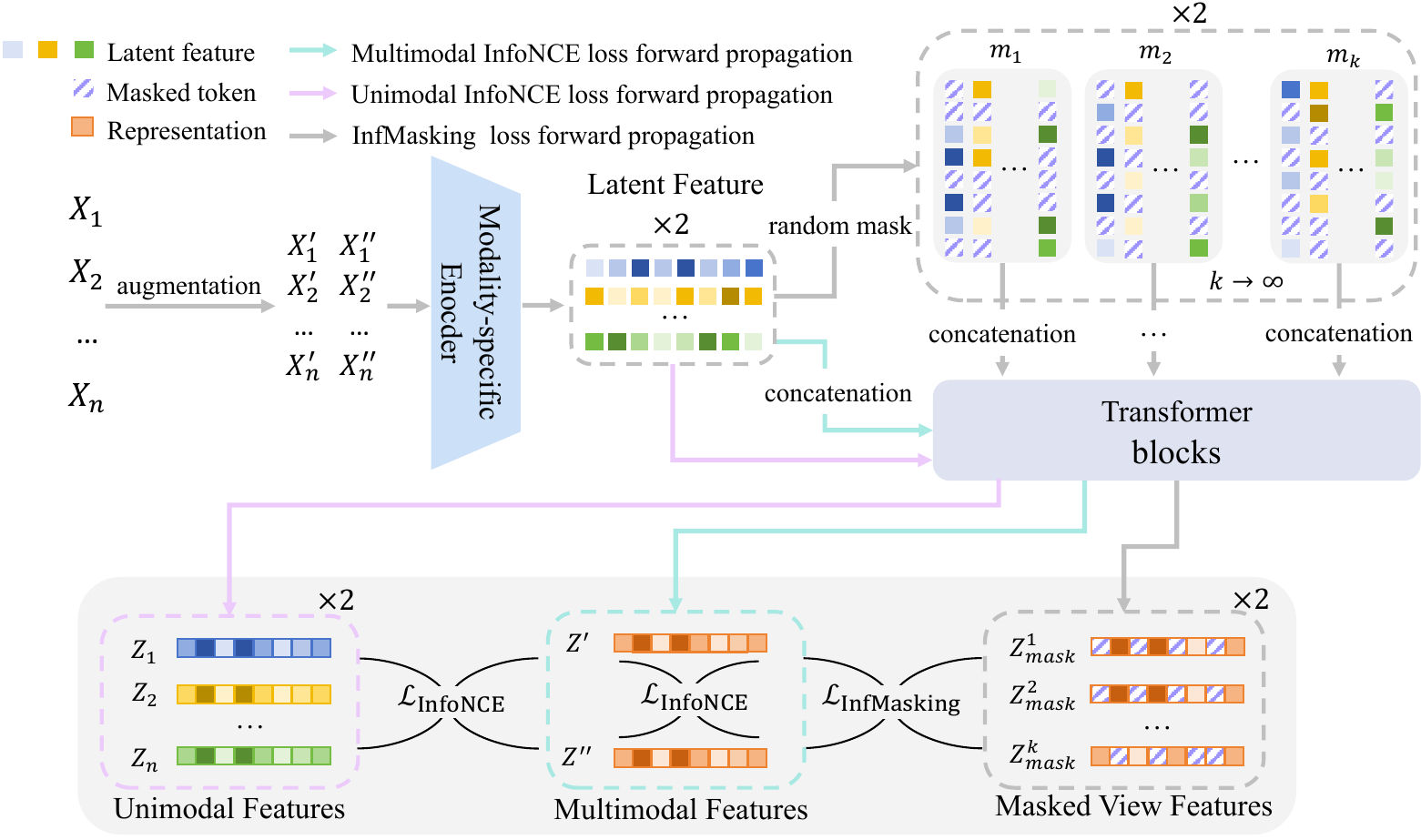}
    \caption{
    The overall pipeline of InfMasking.
    Given \( n \) modalities \( X = (X_1, X_2, \ldots, X_n) \), we augment them to obtain \( X' \) and \( X'' \), which are then encoded independently by modality-specific encoders to extract latent features. These features are processed in three ways:
    (1) All modality features are concatenated and input into a Transformer block, yielding fused features \( Z' \) and \( Z'' \);
    (2) Each modality feature is individually input into a Transformer block, producing unimodal features \( Z_1, Z_2, \ldots, Z_n \) ; 
    (3) Features of each modality are randomly masked, concatenated, and input into a Transformer block, repeated \( k \) times to obtain \( Z^1_{\text{mask}}, Z^2_{\text{mask}}, \ldots, Z^k_{\text{mask}} \).
    }
    \label{fig:flow_chart}
\end{figure}

\subsection{The General Framework }
The proposed framework, termed {\model}, is a multimodal contrastive interaction method designed to enhance synergistic information across modalities by leveraging infinite masked views. The overall pipeline of \model is illustrated in Fig.~\ref{fig:flow_chart} and consists of two primary stages: modality-specific latent feature encoding and multimodal feature fusion via a Transformer.
Given an input set of \( n \) modalities \( X = (X_1, X_2, \ldots, X_n) \), we obtain \( X' \) and \( X'' \) through an augmentation process. Subsequently, \( X' \) and \( X'' \) are processed by modality-specific encoders, where each modality is encoded independently to extract latent features. As shown in Fig.~\ref{fig:flow_chart}, these modality features are then processed in parallel through three distinct ways:
(1) All modality features are concatenated and input into a Transformer block, yielding fused features $Z'$ and $Z''$; (2) Each modality feature is individually input into a Transformer block, producing unimodal features $Z'_1, Z'_2, \ldots, Z'_n$ and $Z''_1, Z''_2, \ldots, Z''_n$; 
(3) Features of each modality are randomly masked, then concatenated and input into a Transformer block to obtain fused features. This process is executed $k$ times, resulting in $Z'^1_{\text{mask}}, Z'^2_{\text{mask}}, \ldots, Z'^k_{\text{mask}}$ and $Z''^1_{\text{mask}}, Z''^2_{\text{mask}}, \ldots, Z''^k_{\text{mask}}$.

Based on Lemma \ref{lemma: Ixxp} and Lemma \ref{cor: Ix1z}, \cite{dufumier2024align} proposes a contrastive multimodal (CoMM) learning  approach to learn task-agnostic multimodal representations by modeling multimodal interactions, including redundancy, uniqueness, and synergy.
CoMM estimates the mutual information using the InfoNCE loss: $\hat{I}_{\text{NCE}}(Z, Z') = \mathbb{E}_{\substack{z,z'_{\text{pos}}\sim p(Z,Z')}} \left[ \log \frac{\exp(z^Tz'_{\text{pos}}/\tau)}{\exp(z^Tz'_{\text{pos}}/\tau)+\sum_{z'_{\text{neg}} } \exp(z^T,z'_{\text{neg}}/\tau)} \right]$, where $\tau$ is a temperature parameter. 
Hence, its training objective is  formulated as follows:
\begin{equation}
    \label{eq:comm-loss}
    \mathcal{L}_\text{CoMM} = -\underbrace{\hat{I}_{\text{NCE}}(Z', Z'')}_{\approx R+S+\sum_{i=1}^n U_i} - \sum_{i=1}^n \underbrace{\tfrac{1}{2}\left(\hat{I}_{\text{NCE}}(Z_i, Z')+ \hat{I}_{\text{NCE}}(Z_i, Z'')\right)}_{\approx R+U_i}.
\end{equation}
While the first term theoretically quantifies redundancy, synergy, and the  unique information across modalities, empirical evidence indicates that its practical performance exhibits notable limitations. Hence, the second term constitutes the fundamental component of CoMM, specifically designed to strengthen both unique and redundant interaction patterns. However, enhancing synergistic interactions remains a substantial challenge in this framework.

Synergy is a complex interaction that arises when different modalities provide complementary information, necessitating their integration for effective task performance. We aim to learn a multimodal representation that captures all three types of interactions, with a particular emphasis on enhancing synergistic information. To achieve this, we introduce \model, a novel approach that leverages infinite masking to enhance the modeling of multimodal interactions. 
Our training objective is  formulated as follows:
\begin{align}
    \label{eq:name-loss}
    \mathcal{L}_\text{Total loss} &= -\underbrace{\hat{I}_{\text{NCE}}(Z', Z'')}_{\approx R + S + \sum_{i=1}^n U_i} - \sum_{i=1}^n \underbrace{\frac{1}{2}\left(\hat{I}_{\text{NCE}}(Z_i, Z') + \hat{I}_{\text{NCE}}(Z_i, Z'')\right)}_{\approx R + U_i} \nonumber\\
    &\quad - \underbrace{\mathbb{E}_{\text{mask}}\left[ \hat{I}_{\text{NCE}}(Z^{\prime}_{\text{mask}}, Z^{\prime}) + \hat{I}_{\text{NCE}}(Z^{\prime\prime}_{\text{mask}}, Z^{\prime\prime}) \right]}_{\mathcal{L}_{\text{InfMasking}}},
\end{align}
where $\mathcal{L}_{\text{InfMasking}}$ represents our novel masking-based regularization term designed to specifically enhance synergistic interactions, as detailed in Section \ref{Contrastive_Synergistic_InfMask}.

\subsection{Contrastive Synergistic Information via Infinite Masking}
\label{Contrastive_Synergistic_InfMask}
In multimodal learning, capturing synergistic information—where different modalities provide complementary insights—is essential for tasks requiring integrated understanding.
We propose a contrastive synergistic Information method via infinte masking to enhance synergistic interactions.  
It's core idea is to randomly mask a significant portion of the features from each modality during the fusion process.
As shown in Fig.~\ref{fig:flow_chart}, we fuse all masked features from different modalities to obtain a fused representation  \(Z^{\prime}_{mask}\) via a Transformer. 
Each time features from each modality are randomly masked, only partial information from each modality is retained. Consequently, after each masking operation,  $Z^{\prime}_{mask}$ contains distinct synergistic information.
Then, by aligning \(Z^{\prime}_{mask}\) with its unmasked counterparts \(Z^{\prime}\)  through maximizing their mutual information, \(Z^{\prime}\) are encouraged to capture distinct synergistic information. 
This process is repeated for $K$ times of masking, we can obtain the final training loss:$\frac{1}{K}\sum_{k=1}^K\hat{I}_{\text{NCE}}(Z^{\prime~k}_{\text{mask}}, Z^{\prime}) +\hat{I}_{\text{NCE}}(Z^{\prime\prime~k}_{\text{mask}}, Z^{\prime\prime})$. 
To enable the model to learn diverse forms of synergistic information, we allow $K$ to approach infinity through infinite masking, ultimately obtaining the masking loss $\mathcal{L}_{\text{\model}}$ as follows:
\begin{align}\label{eq:mask_loss}
    \mathcal{L}_{\text{\model}} &= -\lim _{K\rightarrow \infty}\frac{1}{K}\sum_{k=1}^K\hat{I}_{\text{NCE}}(Z^{\prime~k}_{\text{mask}}, Z^{\prime}) +\hat{I}_{\text{NCE}}(Z^{\prime\prime~k}_{\text{mask}}, Z^{\prime\prime}) \nonumber \\&=  -\mathbb{E}_{\text{mask}}\left[ \hat{I}_{\text{NCE}}(Z^{\prime}_{\text{mask}}, Z^{\prime}) + \hat{I}_{\text{NCE}}(Z^{\prime\prime}_{\text{mask}}, Z^{\prime\prime}) \right].
\end{align}

This infinite masking strategy enables \model to capture richer synergistic interactions by exposing the model to diverse combinations of partial modality information during training. However, the estimation of $    \mathbb{E}_{\text{mask}}\left[ \hat{I}_{\text{NCE}}(Z^{\prime}_{\text{mask}}, Z^{\prime})  \right]$ and $    \mathbb{E}_{\text{mask}}\left[ \hat{I}_{\text{NCE}}(Z^{\prime\prime}_{\text{mask}}, Z^{\prime\prime})  \right]$ is computationally expensive via random mask samples.

To address this issue, we derive a lower bound for $\mathbb{E}_{\text{mask}}\left[ \hat{I}_{\text{NCE}}(Z^{\prime}_{\text{mask}}, Z^{\prime})  \right]$ to optimize the InfMasking loss function Eq.~(\ref{eq:mask_loss}). The detailed derivation is as follows:
\begin{align}
   & \mathbb{E}_{\text{mask}}[\hat{I}_{\text{NCE}}(Z^{\prime}_{\text{mask}}, Z^{\prime})] =\mathbb{E}_{\text{mask}}[\mathbb{E}_{z'\sim p(Z')}\left[ \log \frac{\exp (z'^Tz^{\prime}_{{\text{mask}}}/\tau)}{\exp (z'^Tz^{\prime}_{{\text{mask}}}/\tau)+\sum_{z'_{\text{neg}} } \exp (z^{\prime T}z^{\prime}_{neg}/\tau)} \right] ] \\ &
   =\mathbb{E}_{z'\sim p(Z')}[\mathbb{E}_{\text{mask}}\left[ (z'^Tz^{\prime}_{{\text{mask}}}/\tau)-\log[ \exp (z'^Tz^{\prime}_{{\text{mask}}}/\tau)+\sum_{z'_{\text{neg}} }\exp (z^{\prime T}z^{\prime}_{neg}/\tau)] \right] ]
   \\ &\geq\mathbb{E}_{z'\sim p(Z')}\left[ z^{\prime~T}\mathbb{E}_{\text{mask}}[z^{\prime}_{{\text{mask}}}]/\tau-\log \mathbb{E}_{\text{mask}}[ \exp (z'^Tz^{\prime}_{{\text{mask}}}/\tau)+\sum_{z'_{\text{neg}} }\exp (z^{\prime T}z^{\prime}_{neg}/\tau)] \right ]\label{eq:inequality1}
\end{align}
The inequality Eq.~(\ref{eq:inequality1}) merges from the application of Jensen inequality on concave functions i.e., $\mathbb{E}_x \log (X) \leq \log \mathbb{E}_x[X]$. 
$z^{\prime}_{{\text{mask}}}$ denotes the integrated representation derived from the fusion of all masked features across diverse modalities via the Transformer architecture. 

Inspired by \cite{cai2020joint}, we posit that $z^{\prime}_{{\text{mask}}}$ follows a Gaussian distribution, formally expressed as $z_{\text{mask}}^{\prime} \sim \mathcal{N}(\boldsymbol{\mu}_{z_{\text{mask}}^{\prime}}, \boldsymbol{\Sigma}_{z_{\text{mask}}^{\prime}})$, where $\boldsymbol{\mu}_{z_{\text{mask}}^{\prime}}$ and $\boldsymbol{\Sigma}_{z_{\text{mask}}^{\prime}}$ denote the mean vector and covariance matrix of $z^{\prime}_{{\text{mask}}}$, respectively. This assumption is well-founded for two principal reasons. First, the masked embeddings tend to cluster around a central value in the embedding space, as they all inherently reflect aspects of the query's semantic nature. Second, the variance observed across feature dimensions can be interpreted as a representation of semantic differentiation in the ambient space, which aligns with established principles in distributional semantics. Under this assumption, we can derive:
\begin{lemma}\label{lemma: ifif}
    Let $\boldsymbol{\mu}_{z_{\text{mask}}^{\prime}}$ and $\boldsymbol{\Sigma}_{z_{\text{mask}}^{\prime}}$ be the Gaussian mean vector and covariance matrix of $z^{\prime}_{{\text{mask}}}$, respectively. The lower bound of $\mathbb{E}_{\text{mask}}\left[ \hat{I}_{\text{NCE}}(Z^{\prime}_{\text{mask}}, Z^{\prime})  \right]$ can be obtained as follows: $\mathbb{E}_{\text{mask}}[\hat{I}_{\text{NCE}}(Z^{\prime}_{\text{mask}}, Z^{\prime})]$
    \begin{align}
        \geq\mathbb{E}_{z'\sim p(Z')}\left[ z'^T\boldsymbol{\mu}_{z_{\text{mask}}^{\prime}} /\tau-\log [ \exp (z'^T\boldsymbol{\mu}_{z_{\text{mask}}^{\prime}} /\tau+\frac{z'^T\boldsymbol{\Sigma}_{z_{\text{mask}}^{\prime}}z^{\prime}}{2\tau^2})+\sum_{z'_{\text{neg}} } \exp (z^{\prime T}z^{\prime}_{neg}/\tau)] \right ]\label{eq:inequality2}
     \end{align}  
\end{lemma}
This allows us to approximate the mutual information between the masked and unmasked representations without requiring exhaustive sampling of all possible masks.  A detailed proof is given in Appendix \ref{aprooof}.

\section{Experiments}
\label{experiments}
We perform experiments on both synthetic benchmarks and multiple large-scale real-world datasets to verify the effectiveness of \model in learning representations from diverse modalities.
To evaluate InfMasking's capacity to capture three essential aspects of multimodal interactions (\textit{i.e.}, uniqueness, redundancy, and synergy), we generate synthetic data in a controlled environment based on the Trifeature dataset~\cite{hermann2020shapes}. Furthermore, we assess the generalizability of \model on several widely used multimodal benchmark datasets involving diverse modality combinations in real-world scenarios. These tasks span various domains (\textit{e.g.}, healthcare, robotics, \textit{etc.}) 
allowing for a thorough assessment of the model's representation capabilities across diverse modalities.
Detailed experimental settings are provided in Appendix \ref{Experiment_Details}.
For evaluation, we use linear probing, \textit{i.e.}, freezing the pre-trained feature extractor and training a linear classifier (or regressor, depending on the task) on top of the learned representations. The downstream task performance of the linear model serves as an indicator of the quality of the learned multimodal representations.

\subsection{Synthetic Experiments on Trifeature Datasets}

Following the experimental design of the Trifeature dataset in CoMM~\cite{dufumier2024align}, we conduct controlled experiments on a synthetic dataset derived from Trifeature. 
We assess the model's capacity to learn uniqueness, redundancy and synergy through two separate experiments. 
In terms of uniqueness and redundancy, we define shapes as redundant features and textures as uniqueness features. And the task involves two subtasks: (1) identifying the shared shape between two images (redundancy) and (2) predicting the texture of the first image (or second image) (uniqueness). The random-guessing baselines in both cases corresponds to 10\%. 
As for synergy, we artificially introduce a strong correlation between textures and colors by defining a mapping $\mathcal{M}$ in the training set (\textit{e.g.}, blue maps to triangles, stripes to red), resulting in a 50\% baseline for random guessing. The model is trained on image pairs that follow this mapping. The task is to determine whether a given image pair satisfies the mapping $Y=\mathds{1}(\mathrm{texture}(X_1),\mathrm{color}(X_2)\in\mathcal{M})$, thereby evaluating the model's ability to capture synergistic interactions across modalities. 
\begin{wraptable}{r}{0.68\textwidth}
    \centering
    \vspace{4mm}
    \caption{Linear probing accuracy (in \%) of redundancy (shape), uniqueness (texture) and synergy (color and texture) on Trifeature dataset. $^\clubsuit$ denotes results are from~\cite{dufumier2024align}.}
    \label{tab:trifeature_results}
    \begin{tabular}{lccc}
        \toprule
        \textit{Model} & \textit{redundancy$\uparrow$} & \textit{uniqueness$\uparrow$} & \textit{synergy$\uparrow$} \\
        \midrule
        Cross$^\clubsuit$~\cite{radford2021learning} & $\textbf{100.0}$ & $11.6$ & $50.0$ \\
        Cross+Self$^\clubsuit$~\cite{yuan2021multimodal} & $99.7$ & $86.9$ & $50.0$ \\
        FactorCL$^\clubsuit$~\cite{liang2023factorized} & $99.8$ & $62.5$ & $46.5$ \\
        MAE~\cite{he2022masked} & $99.8_{\pm0.11}$ & $82.4_{\pm3.09}$ & $50.1_{\pm0.24}$ \\
        CoMM~\cite{dufumier2024align} & $99.9_{\pm0.06}$ & $86.8_{\pm2.99}$ & $71.4_{\pm3.47}$ \\
        \model (ours) & $99.9_{\pm0.09}$ & $\textbf{90.6}_{\pm2.31}$ & $\textbf{77.0}_{\pm4.22}$ \\
        \bottomrule
    \end{tabular}
\end{wraptable}

Experimental results are illustrated in Tab.~\ref{tab:trifeature_results}. Cross-modality constraints based on the InfoNCE loss~\cite{radford2021learning} ("Cross" model) achieve the best performance at capturing redundant information but struggle with uniqueness and synergy. FactorCL~\cite{liang2023factorized}, self-supervised constraints on each encoder ("Cross + Self"~\cite{yuan2021multimodal}) and MAE~\cite{he2022masked} (implementation details are provided in Appendix~\ref{MAE_difference}) improve on uniqueness but remain limited in modeling synergy.
CoMM~\cite{dufumier2024align} performs well across all three interactions. However, it still has considerable room for improvement, particularly in capturing synergistic information. In comparison, \model achieves superior performance in capturing both redundancy and synergy, outperforming CoMM by 3.8\% and 5.6\%, respectively. 

\subsection{Experiments on Real-world Datasets}

We further evaluate the performance of our model on several real-world multimodal datasets provided by Multibench~\cite{liang2021multibench}. These datasets span diverse modality combinations and task types, providing a comprehensive benchmark to assess the model's ability to learn effective multimodal representations.
Further dataset details are provided in Appendix~\ref{dataset}.

\subsubsection{ Experiments with 2 Modalities on Multibench}
\label{real_world_dataset}

Following the data preprocessing procedure of previous work~\cite{liang2023factorized,dufumier2024align}, we conduct our experiments using the same encoders, modality configurations and train models based on encoded inputs with diverse modalities. We consider "Cross", "Cross+Self", FactorCL and CoMM as baselines for comparison. 
\begin{table}[ht]
    \centering
    \vspace{-2.5mm}
    \caption{Linear probing MSE($\times 10^{-4}$) for regression task and top-1 accuracy (in \%) for classification tasks on Multibench. $^\clubsuit$ denotes results are from~\cite{liang2023factorized}. $^*$ denotes average is only selected from the results of classification tasks.
    }
    \label{tab:Multibench_results}
    \resizebox{\textwidth}{!}{
    \begin{tabular}{lccccccc}
        \toprule
        \multirow{2}{*}{\textit{Model}} & \multicolumn{1}{c}{\textit{Regression}} & \multicolumn{6}{c}{\textit{Classification}} \\
        \cmidrule(lr){2-2} \cmidrule(lr){3-8}
        & \textit{V\&T EE}$\downarrow$ & \textit{MIMIC}$\uparrow$ & \textit{MOSI}$\uparrow$ & \textit{UR-FUNNY}$\uparrow$ & \textit{MUSTARD}$\uparrow$ & & \textbf{Average}$^{*}\uparrow$\\
        \midrule
        $\text{Cross}^\clubsuit$~\cite{radford2021learning} & $33.09_{\pm3.67}$ & $66.7_{\pm0.1}$ & $47.8_{\pm1.8}$ & $50.1_{\pm1.9}$ & $53.5_{\pm2.9}$ &  & $54.52$ \\
        $\text{Cross+Self}^\clubsuit$~\cite{yuan2021multimodal} & $7.56_{\pm0.31}$ & $65.49_{\pm0.0}$ & $49.0_{\pm1.1}$ & $59.9_{\pm0.9}$ & $53.9_{\pm4.0}$ &  & $57.07$ \\
        $\text{FactorCL}^\clubsuit$~\cite{liang2023factorized} & $10.82_{\pm0.56}$ & $67.3_{\pm0.0}$ & $51.2_{\pm1.6}$ & $60.5_{\pm0.8}$ & $55.80_{\pm0.9}$ &  & $58.7$ \\
        CoMM~\cite{dufumier2024align} & $7.96_{\pm2.13}$ & $66.4_{\pm0.41}$ & $63.7_{\pm2.5}$ & $63.3_{\pm0.51}$ & $64.4_{\pm1.1}$ &  & $64.45$ \\
        \model (ours) & $\textbf{4.23}_{\pm0.37}$ & $\textbf{68.1}_{\pm0.42}$ & $\textbf{69.0}_{\pm1.2}$ & $\textbf{64.3}_{\pm0.9}$ & $\textbf{66.8}_{\pm2.5}$ &  & $\textbf{67.05}$ \\
    \bottomrule
    \end{tabular}
    }
\end{table}
As presented in Tab.~\ref{tab:Multibench_results}, \model consistently achieves the best performance across all benchmark datasets. In the binary classification tasks, \model outperforms CoMM—the strongest baseline—by 1.7\%, 5.3\%, 1.0\%, and 2.4\% on the MIMIC, MOSI, UR-FUNNY, and MUSTARD datasets, respectively. For regression tasks, \model also delivers superior results, achieving a lead of $3 \times 10^{-4}$ in terms of MSE compared to the second-best model. 
These experimental results demonstrate the effectiveness of \model in capturing bimodal interactions. Furthermore, its consistently strong performance across diverse datasets highlights the generalizability and robustness of \model in real-world bimodal scenarios. 

\subsubsection{Experiments with 3 Modalities on Multibench}

\begin{wraptable}{r}{0.57\textwidth}
    \centering
    \vspace{3mm}
    \caption{Linear probing top-1 accuracy (in \%) for classification tasks on Vision\&Touch and UR-FUNNY. $^\clubsuit$ denotes results are from ~\cite{dufumier2024align}.}
    \label{tab:tri_results}
    \small
    \begin{tabular}{lccc}
        \toprule
        \textit{Model} & \textit{\#Mod.} & \textit{V\&T CP$\uparrow$} & \textit{UR-FUNNY$\uparrow$} \\
        \midrule
        Cross & 2 & $86.3_{\pm0.25}$ & $50.1^\clubsuit$ \\
        Cross+Self & 2 & $87.6_{\pm0.26}$ & $59.9^\clubsuit$ \\
        CoMM & 2 & $87.0_{\pm1.77}$ & $63.3_{\pm0.51}$ \\
        \model (ours) & 2 & $88.5_{\pm0.33}$ & $64.3_{\pm0.9}$ \\
        \midrule
        CMC$^\clubsuit$~\cite{tian2020contrastive} & 3 & $94.1$ & $59.2$ \\
        CoMM & 3 & $\textbf{94.1}_{\pm0.17}$ & $64.8_{\pm1.13}$ \\
        \model (ours) & 3 & $94.1_{\pm0.09}$ & $\textbf{65.6}_{\pm1.15}$ \\
        \bottomrule
    \end{tabular}
\end{wraptable}

We evaluate the generalizability of \model in learning multimodal representations beyond two modalities. Specifically, we conduct experiments on two datasets: Vision\&Touch (for the contact prediction task, with visual, force, and proprioceptive modalities) and UR-FUNNY (with visual, text, and audio modalities). CMC~\cite{tian2020contrastive} and CoMM are selected as baselines for comparison in the three-modality setting.

The results are summarized in Tab.~\ref{tab:tri_results}. To more intuitively assess the information gain introduced by incorporating a third modality, we additionally report results from bi-modal training scenarios using CoMM, "Cross" and "Cross + Self". Specifically, we train these baselines on (1) the image and proprioceptive modalities of the Vision\&Touch dataset, and (2) the image and text modalities of the UR-FUNNY dataset.
Our experiments reveal that adding a third modality significantly enhances the performances of CoMM and InfMasking. CoMM as a strong baseline shows performance gains of 7.1\% and 1.5\% on Vision\&Touch and UR-FUNNY, respectively. 
Although InfMasking's performance gain from adding the third modality is relatively modest compared to CoMM, it still matches CoMM's performance on the Vision\&Touch dataset.  On the UR-FUNNY dataset, \model achieves the best result (+0.8\%).

\subsubsection{Experiments with 2 Modalities on Multimodal IMDb}

\begin{wraptable}{r}{0.7\textwidth}
    \centering
    \caption{Linear probing F1-score (weighted and macro) (in \%) for MM-IMDB. $^\triangle$ indicates further training on unlabeded data. $^\clubsuit$ denotes results are from~\cite{dufumier2024align}. }
    \label{tab:imdb_results}
    \small
    \begin{tabular}{lcccc}
        \toprule
        \textit{Model} & \textit{Modalities} & \textit{weighted-f1$\uparrow$} & \textit{macro-f1$\uparrow$} \\
        \midrule
        SimCLR$^{\clubsuit\triangle}$~\cite{chen2020simple} & V & $40.35_{\pm0.23}$ & $27.99_{\pm0.33}$ \\
        \multirow{3}{*}{CLIP$^\clubsuit$~\cite{radford2021learning}} & V & $51.5$ & $40.8$ \\ & L & $51.0$ & $43.0$ \\
        & V+L & $58.9$ & $50.9$ \\
        SLIP$^{\clubsuit\triangle}$~\cite{mu2022slip} & V+L & $56.54_{\pm0.19}$ & $47.35_{\pm0.27}$ \\
        CLIP$^{\clubsuit\triangle}$~\cite{radford2021learning} & V+L & $54.49_{\pm0.19}$ & $44.94_{\pm0.30}$ \\
        CoMM$_{\text{(CLIP backbone)}}$ & V+L & $61.29_{\pm0.73}$ & $53.79_{\pm0.22}$ \\
        \model$_{\text{(ours, CLIP backbone)}}$ & V+L & $\textbf{62.60}_{\pm0.26}$ & $\textbf{55.93}_{\pm0.19}$ \\
        \bottomrule
    \end{tabular}
\end{wraptable}

Multimodal IMDb(MM-IMDb)~\cite{arevalo2017gated} is a real-world multimodal, multi-label dataset designed for movie genre classification. It poses two major challenges: significant class imbalance with genres such as comedy and drama dominating the label distribution, and substantial semantic discrepancy between visual (poster) and textual (plot's description) modalities. Since genre prediction based on a single modality is often unreliable and the combination of both modalities can perform better~\cite{arevalo2017gated}, this underscores the need for effective modeling of multimodal interactions. 
We select both single-modal and multi-modal as baselines. For unimodal, we choose SimCLR (image-only)~\cite{chen2020simple} and CLIP (pretrained on image-text pairs)~\cite{radford2021learning}. For multimodal, we include CLIP, SLIP~\cite{mu2022slip}, and CoMM.

Tab.~\ref{tab:imdb_results} summarizes the experimental results. Models trained on both modalities consistently outperformed their single-modality counterparts, further validating the importance of optimizing multimodal representation learning. \model achieves the best overall performance, improving upon CoMM by 1.31\% in weighted F1-score and 2.14\% in macro F1-score. It is also worth noting that CLIP with its original public weights achieves 58.9\% on weighted F1-score, outperforming CLIP fine-tuned on MM-IMDb (54.59\%). This suggests that redundant information learning may not always benefit complex tasks such as genre prediction, which require complementary modality alignment~\cite{dufumier2024align}. These results demonstrate the robustness and generalizability of \model in handling imbalanced, semantically heterogeneous, and multi-label multimodal classification tasks.

\section{Ablation Studies}

To examine the effectiveness of the design of InfMasking, we conduct comprehensive ablation studies on the bimodal Trifeature dataset focusing on three critical components: the loss function formulation, the optimal number of masked views, and the masking ratio parameter. 

\begin{table}[h]
    \centering
    \vspace{-2.5mm}
    \caption{Linear probing accuracy (in \%) of redundancy \textit{R}, uniqueness \textit{U} and synergy \textit{S} on Trifeature Dataset under different combinations of loss functions. $\lambda_1$, $\lambda_2$, and $\lambda_3$ denote the weights for $\mathcal{L}_{mask}$, $\sum_{i} \mathcal{L}_{i}$, and $\mathcal{L}$, respectively, where $\mathcal{L}$ and $\sum_{i} \mathcal{L}_{i}$ correspond to the first and second terms in Eq.~\ref{eq:comm-loss}. }
    \label{tab:loss_weights}
    \begin{tabular}{cccccccc}
        \toprule
        \multicolumn{3}{c}{\textit{loss weights}}  & \multirow{2}{*}{\textit{R}} & \multirow{2}{*}{\textit{$U_1$}} & \multirow{2}{*}{\textit{$U_2$}} & \multirow{2}{*}{\textit{S}} & \multirow{2}{*}{\textit{Average}} \\
        \cmidrule(lr){1-3} 
        \(\lambda_1\) & \(\lambda_2\) & \(\lambda_3\)  \\
        \midrule
        0 & 0 & 1 & $95.8_{\pm1.91}$ & $85.9_{\pm2.11}$ & $83.8_{\pm2.97}$ & $58.7_{\pm7.11}$ & $80.1$ \\
        0 & 1 & 1 & $99.9_{\pm0.06}$ & $87.1_{\pm3.31}$ & $86.5_{\pm2.60}$ & $71.4_{\pm3.47}$ & $86.0$\\
        1 & 1 & 0 & $\textbf{99.9}_{\pm0.08}$ & $\textbf{90.7}_{\pm2.10}$ & $\textbf{91.4}_{\pm3.03}$ & $69.2_{\pm6.20}$ & $87.8$ \\
        1 & 1 & 1 & $99.9_{\pm0.09}$ & $90.3_{\pm1.52}$ & $90.8_{\pm2.88}$ & $\textbf{77.0}_{\pm4.22}$ & $\textbf{89.5}$ \\
        \bottomrule
        \end{tabular}
\end{table}

\textbf{Loss function.}
We conducted an ablation study on the Trifeature dataset to evaluate different loss combinations for capturing multimodal interactions. As shown in Tab.~\ref{tab:loss_weights}, the full objective ($\lambda_1 = \lambda_2 = \lambda_3 = 1$, InfMasking) achieves the highest synergy at 77.0\% while maintaining balanced performance across other metrics.
Using only CoMM loss ($\lambda_1 = 0$, $\lambda_2 = 1$, $\lambda_3 = 1$) yields 71.4\% synergy, while using only $\mathcal{L}$ ($\lambda_1 = 0$, $\lambda_2 = 0$, $\lambda_3 = 1$) further decreases to 58.7\%, indicating that CoMM loss alone is insufficient without the view diversity from masking.
Excluding the $\mathcal{L}$(\textit{i.e.}, $\hat{I}_{\text{NCE}}(Z', Z'')$) drops synergy to 69.2\%, despite marginal improvements in redundancy and uniqueness scores. As noted in CoMM~\cite{dufumier2024align}, minimizing $\mathcal{L}$ enables the model to capture all information terms, albeit at a slower rate. When this loss is removed, the model learns redundancy and uniqueness more efficiently, achieving higher scores within the same epoch, but at the cost of diminished synergy performance.

\begin{figure}[h]
	\centering
	\subfigure[Number of masked views.]{\label{fig:number}
		\includegraphics[width=0.49\textwidth]{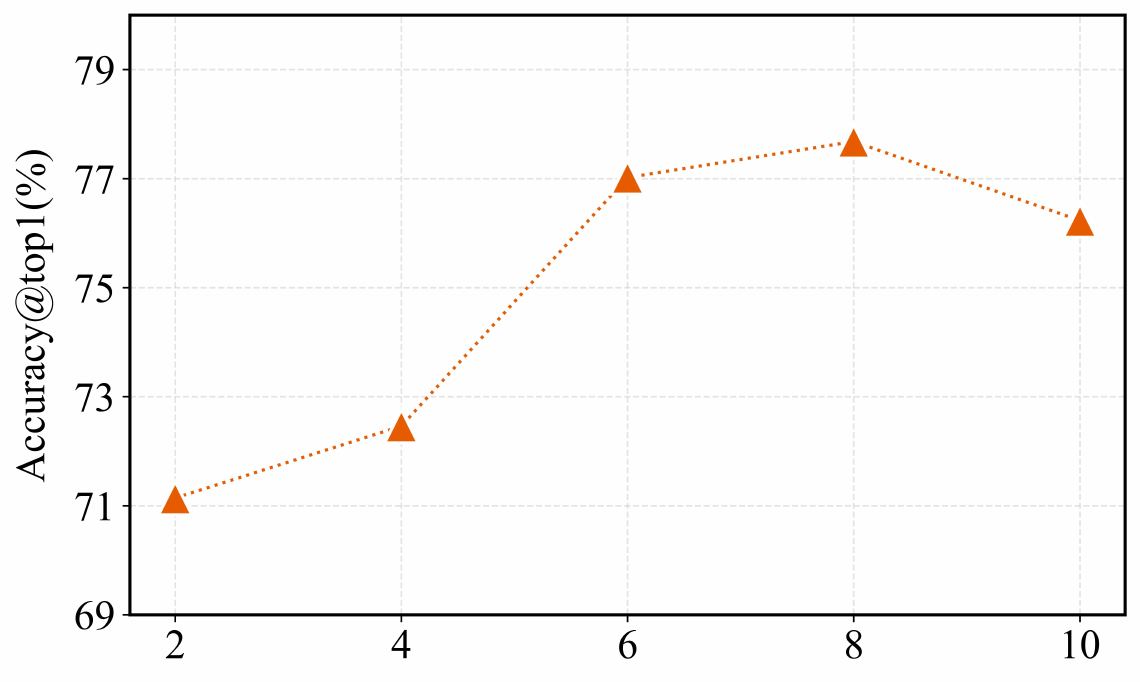}}
	\subfigure[Ratio of masking (\%).]{\label{fig:ratio}
		\includegraphics[width=0.49\textwidth]{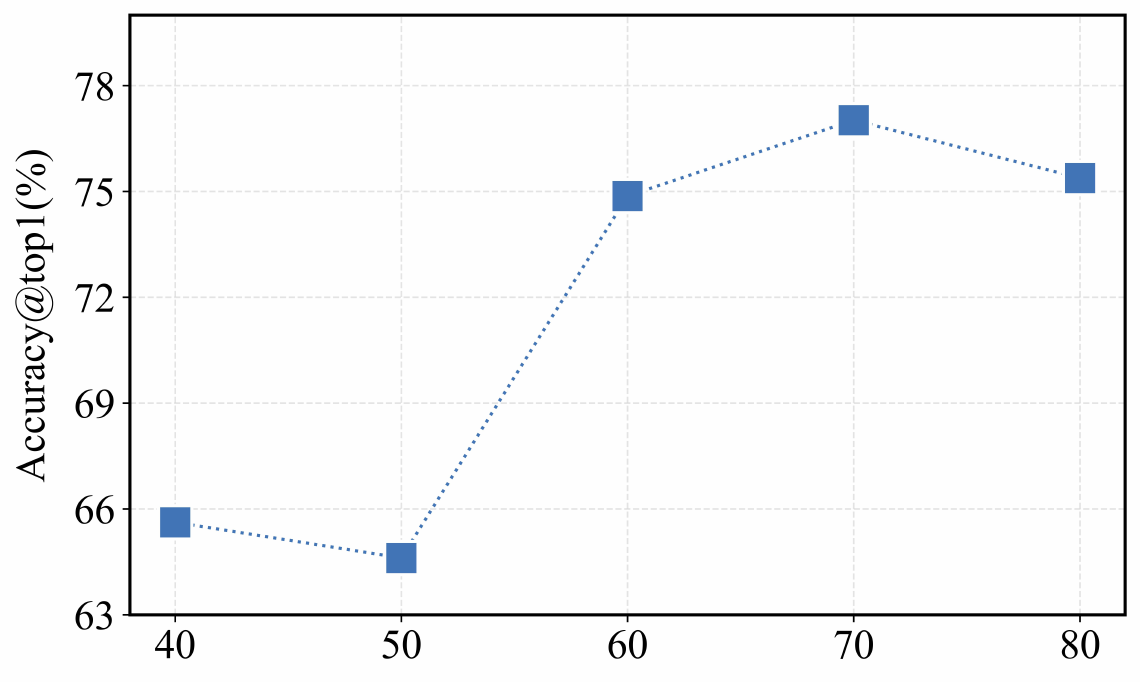}}
	\caption{Synergy accuracy changes with different masked setting on Trifeature datasets.}
	\label{fig:ablation}
\end{figure}

\textbf{Number of masked views.} 
According to Section~\ref{Contrastive_Synergistic_InfMask}, increasing the number of masked views yields a closer approximation to $\mathbb{E}_{\text{mask}}\left[ \hat{I}_{\text{NCE}}(Z^{\prime}_{\text{mask}}, Z^{\prime}) \right]$ and $\mathbb{E}_{\text{mask}}\left[ \hat{I}_{\text{NCE}}(Z^{\prime\prime}_{\text{mask}}, Z^{\prime\prime}) \right]$, albeit at a higher computational cost. As observed in Fig.~\ref{fig:number}, the synergy score improves progressively with an increasing number of views. Notably, performance is sufficiently robust when the number is in the range of $[6, 10]$, demonstrating practical feasibility for GPU implementation. 

\textbf{Masking ratio.} 
Fig.~\ref{fig:ratio} illustrates the impact of varying the masking ratio. At lower ratios ($\leq 50\%$), although the model can capture synergistic information, the overall performance remains unsatisfactory. In contrast, maintaining a higher masking ratio enables the model to generate superior multimodal representations that effectively leverage complementary information across modalities. Furthermore, a higher masking ratio can also provide a greater speedup benefit~\cite{he2022masked}.

\section{Related Work}
\label{related_work}
\textbf{Multimodal learning.}
Multimodal learning integrates diverse data sources—such as text, image, audio, and tactile inputs—to enhance information understanding across modalities~\cite{liang2022foundations,ngiam2011multimodal,guo2025disentangling,liu2025nexus,dai2022one}. Traditional approaches rely on simple fusion techniques like feature concatenation~\cite{d2015review} or modality-specific prediction averaging~\cite{friedman2008predictive}. Transformer architectures revolutionized this field through dynamic cross-modal attention mechanisms~\cite{tsai2019multimodal,xu2023multimodal}.
Contemporary approaches typically follow a two-stage framework: training specialized encoders for each modality, then projecting these representations into a unified embedding space~\cite{baltruvsaitis2018multimodal,dufumier2024align}. This paradigm has been applied across representation learning~\cite{barnum2020benefits,radford2021learning}, cross-modal alignment~\cite{li2022clip,hendricks2021decoupling}, and generative modeling~\cite{ahuja2020style,ramesh2021zero}. 

\textbf{Self-Supervised multimodal representation learning.}
Self-supervised learning generates supervision signals from data's inherent structure~\cite{ericsson2022self,jing2020self,DBLP:conf/iclr/WenZHZX20,DBLP:journals/pami/WenWLX24}. In multimodal contexts, these approaches leverage cross-modal correspondences~\cite{zong2023self, wang2025cooperation}. Prior methods have explored generative approaches, such as reconstructing one modality from another~\cite{ramesh2021zero}, and masked prediction strategies for joint modality modeling~\cite{bachmann2022multimae}.
Contrastive learning has emerged as particularly effective for multimodal representation learning~\cite{girdhar2023imagebind,radford2021learning}, constructing positive pairs through data augmentation~\cite{chen2020simple} and introducing both intra-modal and cross-modal objectives~\cite{mi2024congeo}. Some approaches incorporate regularization terms to align representations across modalities~\cite{wang2022rethinking}.

\textbf{Contrastive Multimodal Interactions.}
Contrastive multimodal approaches~\cite{radford2021learning,jia2021scaling} optimize cross-modal contrastive loss but emphasize redundant information while neglecting unique and synergistic information requiring joint consideration, with FactorCL~\cite{liang2023factorized} addressing this through explicit modeling despite practical implementation challenges.
CoMM~\cite{dufumier2024align} advances the field using multimodal augmentations and information theory-grounded losses to capture various interaction patterns, though enhancing synergistic interactions remains challenging.

\section{Conclusion}
This paper introduces InfMasking, a contrastive method that effectively captures synergistic information in multimodal representation learning by stochastically occluding features during fusion and aligning representations through mutual information maximization. We derive a computationally efficient approximation for infinite masking patterns and demonstrate that our approach not only enhances synergistic information extraction in controlled settings but also achieves state-of-the-art performance across seven diverse multimodal benchmarks. 
\model has some limitations.  It lacks a rigorous theoretical framework to systematically analyze the mechanisms of synergistic interactions. Future research will prioritize developing comprehensive theoretical foundations to formally characterize and measure the synergistic information.

{\large\textbf{Acknowledgments}}

This work was supported by the National Natural Science Foundation of China (No. U24A20323),  the Major Science and Technology Special Project of the Sichuan Provincial Department of Science and Technology (Grant No. 2024ZDZX0002), the Sichuan Provincial Innovation Group Project (Grant No. 2024NSFTD0054), Fundamental Research Funds for the Central Universities (JBK202511081), Blockchain Research Center of China,
Natural Science Foundation of China under Grant No. 62502397,  National Natural Science Foundation of China (NSFC) (72471197), and the Sichuan Provincial Philosophy and Social Science Fund (Grant No. SCJJ25ND091) .
We thank anonymous reviewers for their insightful feedback and Benoit Dufumier, the first author of CoMM~\cite{dufumier2024align}, for his contributions to open sourcing the code.

\newpage
\bibliography{reference.bib}
\bibliographystyle{plain}


\newpage
\section*{NeurIPS Paper Checklist}

\begin{enumerate}

\item {\bf Claims}
    \item[] Question: Do the main claims made in the abstract and introduction accurately reflect the paper's contributions and scope?
    \item[] Answer: \answerYes{} 
    \item[] Justification: We mainly claim 'synergistic information constitutes the fundamental value proposition of multimodal representation', and we propose a contrastive synergistic information method designed to enhance synergistic information through an Infinite strategy. To verify the effectiveness, we tested on real-world datasets across diverse domains (healthcare, robotics) and data types (image, text, audio). InfMasking achieves state-of-the-art results on seven multimodal tasks involving two or three modalities.
    \item[] Guidelines:
    \begin{itemize}
        \item The answer NA means that the abstract and introduction do not include the claims made in the paper.
        \item The abstract and/or introduction should clearly state the claims made, including the contributions made in the paper and important assumptions and limitations. A No or NA answer to this question will not be perceived well by the reviewers. 
        \item The claims made should match theoretical and experimental results, and reflect how much the results can be expected to generalize to other settings. 
        \item It is fine to include aspirational goals as motivation as long as it is clear that these goals are not attained by the paper. 
    \end{itemize}

\item {\bf Limitations}
    \item[] Question: Does the paper discuss the limitations of the work performed by the authors?
    \item[] Answer: \answerYes{} 
    \item[] Justification: We claim the need for further application to pertaining and post-training, which will benefit more, which is presented in the conclusion section.
    \item[] Guidelines:
    \begin{itemize}
        \item The answer NA means that the paper has no limitation while the answer No means that the paper has limitations, but those are not discussed in the paper. 
        \item The authors are encouraged to create a separate "Limitations" section in their paper.
        \item The paper should point out any strong assumptions and how robust the results are to violations of these assumptions (e.g., independence assumptions, noiseless settings, model well-specification, asymptotic approximations only holding locally). The authors should reflect on how these assumptions might be violated in practice and what the implications would be.
        \item The authors should reflect on the scope of the claims made, e.g., if the approach was only tested on a few datasets or with a few runs. In general, empirical results often depend on implicit assumptions, which should be articulated.
        \item The authors should reflect on the factors that influence the performance of the approach. For example, a facial recognition algorithm may perform poorly when image resolution is low or images are taken in low lighting. Or a speech-to-text system might not be used reliably to provide closed captions for online lectures because it fails to handle technical jargon.
        \item The authors should discuss the computational efficiency of the proposed algorithms and how they scale with dataset size.
        \item If applicable, the authors should discuss possible limitations of their approach to address problems of privacy and fairness.
        \item While the authors might fear that complete honesty about limitations might be used by reviewers as grounds for rejection, a worse outcome might be that reviewers discover limitations that aren't acknowledged in the paper. The authors should use their best judgment and recognize that individual actions in favor of transparency play an important role in developing norms that preserve the integrity of the community. Reviewers will be specifically instructed to not penalize honesty concerning limitations.
    \end{itemize}

\item {\bf Theory assumptions and proofs}
    \item[] Question: For each theoretical result, does the paper provide the full set of assumptions and a complete (and correct) proof?
    \item[] Answer: \answerYes{} 
    \item[] Justification: All theorems, formulas, and proofs in the main text are properly numbered and cross-referenced. The key proofs have been included in the appendix~\ref{aprooof} for clarity and completeness.
    \item[] Guidelines:
    \begin{itemize}
        \item The answer NA means that the paper does not include theoretical results. 
        \item All the theorems, formulas, and proofs in the paper should be numbered and cross-referenced.
        \item All assumptions should be clearly stated or referenced in the statement of any theorems.
        \item The proofs can either appear in the main paper or the supplemental material, but if they appear in the supplemental material, the authors are encouraged to provide a short proof sketch to provide intuition. 
        \item Inversely, any informal proof provided in the core of the paper should be complemented by formal proofs provided in appendix or supplemental material.
        \item Theorems and Lemmas that the proof relies upon should be properly referenced. 
    \end{itemize}

    \item {\bf Experimental result reproducibility}
    \item[] Question: Does the paper fully disclose all the information needed to reproduce the main experimental results of the paper to the extent that it affects the main claims and/or conclusions of the paper (regardless of whether the code and data are provided or not)?
    \item[] Answer: \answerYes{} 
    \item[] Justification: This study introduces \textbf{\model}, a multimodal contrastive interaction method that enhances synergistic Information by utilizing Infinite Masking. \model strategically masks substantial portions of features from each modality during the fusion process, and then maximizes the mutual information between these masked fused features and their unmasked counterparts. 
    Through controlled experiments, we demonstrate that \model effectively enhances synergistic information between modalities. In evaluations on large-scale real-world datasets, \model achieves state-of-the-art performance across seven benchmarks. Detailed analysis and results can be found in Sec.~\ref{experiments}. We also present the key hyperparameters in Appendix \ref{Experiment_Details}. 
    \item[] Guidelines:
    \begin{itemize}
        \item The answer NA means that the paper does not include experiments.
        \item If the paper includes experiments, a No answer to this question will not be perceived well by the reviewers: Making the paper reproducible is important, regardless of whether the code and data are provided or not.
        \item If the contribution is a dataset and/or model, the authors should describe the steps taken to make their results reproducible or verifiable. 
        \item Depending on the contribution, reproducibility can be accomplished in various ways. For example, if the contribution is a novel architecture, describing the architecture fully might suffice, or if the contribution is a specific model and empirical evaluation, it may be necessary to either make it possible for others to replicate the model with the same dataset, or provide access to the model. In general. releasing code and data is often one good way to accomplish this, but reproducibility can also be provided via detailed instructions for how to replicate the results, access to a hosted model (e.g., in the case of a large language model), releasing of a model checkpoint, or other means that are appropriate to the research performed.
        \item While NeurIPS does not require releasing code, the conference does require all submissions to provide some reasonable avenue for reproducibility, which may depend on the nature of the contribution. For example
        \begin{enumerate}
            \item If the contribution is primarily a new algorithm, the paper should make it clear how to reproduce that algorithm.
            \item If the contribution is primarily a new model architecture, the paper should describe the architecture clearly and fully.
            \item If the contribution is a new model (e.g., a large language model), then there should either be a way to access this model for reproducing the results or a way to reproduce the model (e.g., with an open-source dataset or instructions for how to construct the dataset).
            \item We recognize that reproducibility may be tricky in some cases, in which case authors are welcome to describe the particular way they provide for reproducibility. In the case of closed-source models, it may be that access to the model is limited in some way (e.g., to registered users), but it should be possible for other researchers to have some path to reproducing or verifying the results.
        \end{enumerate}
    \end{itemize}

\item {\bf Open access to data and code}
    \item[] Question: Does the paper provide open access to the data and code, with sufficient instructions to faithfully reproduce the main experimental results, as described in supplemental material?
    \item[] Answer: \answerYes{} 
    \item[] Justification: After the anonymity period, we will open-source our training code and training data.
    \item[] Guidelines:
    \begin{itemize}
        \item The answer NA means that paper does not include experiments requiring code.
        \item Please see the NeurIPS code and data submission guidelines (\url{https://nips.cc/public/guides/CodeSubmissionPolicy}) for more details.
        \item While we encourage the release of code and data, we understand that this might not be possible, so “No” is an acceptable answer. Papers cannot be rejected simply for not including code, unless this is central to the contribution (e.g., for a new open-source benchmark).
        \item The instructions should contain the exact command and environment needed to run to reproduce the results. See the NeurIPS code and data submission guidelines (\url{https://nips.cc/public/guides/CodeSubmissionPolicy}) for more details.
        \item The authors should provide instructions on data access and preparation, including how to access the raw data, preprocessed data, intermediate data, and generated data, etc.
        \item The authors should provide scripts to reproduce all experimental results for the new proposed method and baselines. If only a subset of experiments are reproducible, they should state which ones are omitted from the script and why.
        \item At submission time, to preserve anonymity, the authors should release anonymized versions (if applicable).
        \item Providing as much information as possible in supplemental material (appended to the paper) is recommended, but including URLs to data and code is permitted.
    \end{itemize}

\item {\bf Experimental setting/details}
    \item[] Question: Does the paper specify all the training and test details (e.g., data splits, hyperparameters, how they were chosen, type of optimizer, etc.) necessary to understand the results?
    \item[] Answer: \answerYes{} 
    \item[] Justification: As shown in Appendix A, we provide comprehensive details of both training and evaluation procedures.
    \item[] Guidelines:
    \begin{itemize}
        \item The answer NA means that the paper does not include experiments.
        \item The experimental setting should be presented in the core of the paper to a level of detail that is necessary to appreciate the results and make sense of them.
        \item The full details can be provided either with the code, in appendix, or as supplemental material.
    \end{itemize}

\item {\bf Experiment statistical significance}
    \item[] Question: Does the paper report error bars suitably and correctly defined or other appropriate information about the statistical significance of the experiments?
    \item[] Answer: \answerYes{} 
    \item[] Justification: For all experiments, we report the mean and standard deviation across five independent runs with random seeds in the range $[42, 46]$.
    \item[] Guidelines:
    \begin{itemize}
        \item The answer NA means that the paper does not include experiments.
        \item The authors should answer "Yes" if the results are accompanied by error bars, confidence intervals, or statistical significance tests, at least for the experiments that support the main claims of the paper.
        \item The factors of variability that the error bars are capturing should be clearly stated (for example, train/test split, initialization, random drawing of some parameter, or overall run with given experimental conditions).
        \item The method for calculating the error bars should be explained (closed form formula, call to a library function, bootstrap, etc.)
        \item The assumptions made should be given (e.g., Normally distributed errors).
        \item It should be clear whether the error bar is the standard deviation or the standard error of the mean.
        \item It is OK to report 1-sigma error bars, but one should state it. The authors should preferably report a 2-sigma error bar than state that they have a 96\% CI, if the hypothesis of Normality of errors is not verified.
        \item For asymmetric distributions, the authors should be careful not to show in tables or figures symmetric error bars that would yield results that are out of range (e.g. negative error rates).
        \item If error bars are reported in tables or plots, The authors should explain in the text how they were calculated and reference the corresponding figures or tables in the text.
    \end{itemize}

\item {\bf Experiments compute resources}
    \item[] Question: For each experiment, does the paper provide sufficient information on the computer resources (type of compute workers, memory, time of execution) needed to reproduce the experiments?
    \item[] Answer: \answerYes{} 
    \item[] Justification: Appendix~\ref{Experiment_Details} provides detailed information about the computational resources utilized in our experiments. 
    \item[] Guidelines:
    \begin{itemize}
        \item The answer NA means that the paper does not include experiments.
        \item The paper should indicate the type of compute workers CPU or GPU, internal cluster, or cloud provider, including relevant memory and storage.
        \item The paper should provide the amount of compute required for each of the individual experimental runs as well as estimate the total compute. 
        \item The paper should disclose whether the full research project required more compute than the experiments reported in the paper (e.g., preliminary or failed experiments that didn't make it into the paper). 
    \end{itemize}
    
\item {\bf Code of ethics}
    \item[] Question: Does the research conducted in the paper conform, in every respect, with the NeurIPS Code of Ethics \url{https://neurips.cc/public/EthicsGuidelines}?
    \item[] Answer: \answerYes{} 
    \item[] Justification: This study develops a representation learning model (InfMasking) using publicly available datasets (compliant with their respective licenses) and does not involve human participants or personally identifiable information, thus requiring no IRB approval. We have provided detailed experimental descriptions and plan to open-source the training code and data after the anonymity period to ensure reproducibility. The paper discusses the method’s limitations and potential societal impacts, including bias risks and mitigation strategies. However, due to the absence of a formal IRB process in the research location, we conducted an internal peer review to address ethical considerations, as suggested by the NeurIPS guidelines. Additionally, local data privacy regulations prevent immediate data sharing during the anonymity period, but we commit to full transparency post-anonymity. After reviewing the NeurIPS Code of Ethics, we confirm that the study complies with most requirements, with the noted exceptions being addressed through alternative measures.
    \item[] Guidelines:
    \begin{itemize}
        \item The answer NA means that the authors have not reviewed the NeurIPS Code of Ethics.
        \item If the authors answer No, they should explain the special circumstances that require a deviation from the Code of Ethics.
        \item The authors should make sure to preserve anonymity (e.g., if there is a special consideration due to laws or regulations in their jurisdiction).
    \end{itemize}

\item {\bf Broader impacts}
    \item[] Question: Does the paper discuss both potential positive societal impacts and negative societal impacts of the work performed?
    \item[] Answer: \answerYes{} 
    \item[] Justification: Please refer to Appendix~\ref{broader_impact}. 
    \item[] Guidelines:
    \begin{itemize}
        \item The answer NA means that there is no societal impact of the work performed.
        \item If the authors answer NA or No, they should explain why their work has no societal impact or why the paper does not address societal impact.
        \item Examples of negative societal impacts include potential malicious or unintended uses (e.g., disinformation, generating fake profiles, surveillance), fairness considerations (e.g., deployment of technologies that could make decisions that unfairly impact specific groups), privacy considerations, and security considerations.
        \item The conference expects that many papers will be foundational research and not tied to particular applications, let alone deployments. However, if there is a direct path to any negative applications, the authors should point it out. For example, it is legitimate to point out that an improvement in the quality of generative models could be used to generate deepfakes for disinformation. On the other hand, it is not needed to point out that a generic algorithm for optimizing neural networks could enable people to train models that generate Deepfakes faster.
        \item The authors should consider possible harms that could arise when the technology is being used as intended and functioning correctly, harms that could arise when the technology is being used as intended but gives incorrect results, and harms following from (intentional or unintentional) misuse of the technology.
        \item If there are negative societal impacts, the authors could also discuss possible mitigation strategies (e.g., gated release of models, providing defenses in addition to attacks, mechanisms for monitoring misuse, mechanisms to monitor how a system learns from feedback over time, improving the efficiency and accessibility of ML).
    \end{itemize}
    
\item {\bf Safeguards}
    \item[] Question: Does the paper describe safeguards that have been put in place for responsible release of data or models that have a high risk for misuse (e.g., pretrained language models, image generators, or scraped datasets)?
    \item[] Answer: \answerNA{} 
    \item[] Justification: We utilize the pre-processed datasets provided by MultiBench~\cite{liang2021multibench}, which have been anonymized to safeguard personal privacy. For the MM-IMDb experiments, we employ publicly available open-source pre-trained models that adhere to established security guidelines.
    \item[] Guidelines:
    \begin{itemize}
        \item The answer NA means that the paper poses no such risks.
        \item Released models that have a high risk for misuse or dual-use should be released with necessary safeguards to allow for controlled use of the model, for example by requiring that users adhere to usage guidelines or restrictions to access the model or implementing safety filters. 
        \item Datasets that have been scraped from the Internet could pose safety risks. The authors should describe how they avoided releasing unsafe images.
        \item We recognize that providing effective safeguards is challenging, and many papers do not require this, but we encourage authors to take this into account and make a best faith effort.
    \end{itemize}

\item {\bf Licenses for existing assets}
    \item[] Question: Are the creators or original owners of assets (e.g., code, data, models), used in the paper, properly credited and are the license and terms of use explicitly mentioned and properly respected?
    \item[] Answer: \answerYes{} 
    \item[] Justification: We have provided detailed descriptions and clearly marked the sources and citations for all models and frameworks involved in the paper within the experimental section. For open-source code, we have included comprehensive comments and explanations for all imported packages and foundational code used.
    \item[] Guidelines:
    \begin{itemize}
        \item The answer NA means that the paper does not use existing assets.
        \item The authors should cite the original paper that produced the code package or dataset.
        \item The authors should state which version of the asset is used and, if possible, include a URL.
        \item The name of the license (e.g., CC-BY 4.0) should be included for each asset.
        \item For scraped data from a particular source (e.g., website), the copyright and terms of service of that source should be provided.
        \item If assets are released, the license, copyright information, and terms of use in the package should be provided. For popular datasets, \url{paperswithcode.com/datasets} has curated licenses for some datasets. Their licensing guide can help determine the license of a dataset.
        \item For existing datasets that are re-packaged, both the original license and the license of the derived asset (if it has changed) should be provided.
        \item If this information is not available online, the authors are encouraged to reach out to the asset's creators.
    \end{itemize}

\item {\bf New assets}
    \item[] Question: Are new assets introduced in the paper well documented and is the documentation provided alongside the assets?
    \item[] Answer: \answerYes{} 
    \item[] Justification: The paper introduces new assets, including the training code and dataset used for the InfMasking representation learning model. These assets are thoroughly documented in the experimental section, which provides detailed descriptions of the model architecture, training procedures, and dataset characteristics, along with citations to all foundational frameworks and packages used. For the open-source code, we include comprehensive comments explaining the functionality of imported packages and the structure of the foundational code. A structured documentation template, detailing training configurations, dataset licensing (compliant with applicable open-source licenses), and model limitations, will be provided alongside the assets. 
    \item[] Guidelines:
    \begin{itemize}
        \item The answer NA means that the paper does not release new assets.
        \item Researchers should communicate the details of the dataset/code/model as part of their submissions via structured templates. This includes details about training, license, limitations, etc. 
        \item The paper should discuss whether and how consent was obtained from people whose asset is used.
        \item At submission time, remember to anonymize your assets (if applicable). You can either create an anonymized URL or include an anonymized zip file.
    \end{itemize}

\item {\bf Crowdsourcing and research with human subjects}
    \item[] Question: For crowdsourcing experiments and research with human subjects, does the paper include the full text of instructions given to participants and screenshots, if applicable, as well as details about compensation (if any)? 
    \item[] Answer: \answerNA{} 
    \item[] Justification: This work does not involve crowdsourcing nor research with human subjects. 
    \item[] Guidelines:
    \begin{itemize}
        \item The answer NA means that the paper does not involve crowdsourcing nor research with human subjects.
        \item Including this information in the supplemental material is fine, but if the main contribution of the paper involves human subjects, then as much detail as possible should be included in the main paper. 
        \item According to the NeurIPS Code of Ethics, workers involved in data collection, curation, or other labor should be paid at least the minimum wage in the country of the data collector. 
    \end{itemize}

\item {\bf Institutional review board (IRB) approvals or equivalent for research with human subjects}
    \item[] Question: Does the paper describe potential risks incurred by study participants, whether such risks were disclosed to the subjects, and whether Institutional Review Board (IRB) approvals (or an equivalent approval/review based on the requirements of your country or institution) were obtained?
    \item[] Answer: \answerNA{} 
    \item[] Justification: This work does not involve crowdsourcing nor research with human subjects. 
    \item[] Guidelines:
    \begin{itemize}
        \item The answer NA means that the paper does not involve crowdsourcing nor research with human subjects.
        \item Depending on the country in which research is conducted, IRB approval (or equivalent) may be required for any human subjects research. If you obtained IRB approval, you should clearly state this in the paper. 
        \item We recognize that the procedures for this may vary significantly between institutions and locations, and we expect authors to adhere to the NeurIPS Code of Ethics and the guidelines for their institution. 
        \item For initial submissions, do not include any information that would break anonymity (if applicable), such as the institution conducting the review.
    \end{itemize}

\item {\bf Declaration of LLM usage}
    \item[] Question: Does the paper describe the usage of LLMs if it is an important, original, or non-standard component of the core methods in this research? Note that if the LLM is used only for writing, editing, or formatting purposes and does not impact the core methodology, scientific rigorousness, or originality of the research, declaration is not required.
    \item[] Answer: \answerNA{} 
    \item[] Justification: In this work, the LLM is utilized solely for writing and editing purposes. They do not influence the core methodology, scientific rigor, or the originality of the research.
    \item[] Guidelines:
    \begin{itemize}
        \item The answer NA means that the core method development in this research does not involve LLMs as any important, original, or non-standard components.
        \item Please refer to our LLM policy (\url{https://neurips.cc/Conferences/2025/LLM}) for what should or should not be described.
    \end{itemize}

\end{enumerate}

\newpage

\appendix
\newcommand{\reddashedline}{\textcolor{red}{\rule[0.5ex]{0.3em}{1pt}\hspace{0.2em}\rule[0.5ex]{0.3em}{1pt}\hspace{0.2em}\rule[0.5ex]{0.3em}{1pt}\hspace{0.2em}\rule[0.5ex]{0.3em}{1pt}}}

\begin{center}
    \LARGE \textbf{Appendix}
\end{center}

\section{Experimental Details}
\label{Experiment_Details}

\textbf{Training protocol. }
All experiments are conducted using five independent runs with random seeds in the range $[42, 46]$. We report the mean and standard deviation of performance metrics (\textit{i.e.}, accuracy, mean squared error) to account for variability across runs. Early stopping based on validation accuracy is systematically applied to prevent overfitting. The best-performing checkpoint on the validation set is selected for final evaluation on the test set.

For dataset-specific encoder architectures, modality-specific data augmentation and latent converters, we follow the same configurations as CoMM~\cite{dufumier2024align}.

\textbf{Training details. }
We use AdamW~\cite{loshchilov2017decoupled} as the optimizer in all experiments. Detailed hyperparameters are listed in Tab.~\ref{tab:hyperparameters}. 
Following~\cite{dufumier2024align} on MM-IMDb, we also use a cosine scheduler with final value $10^{-6}$ and a warmup over 10 epochs. And all models are trained for 100 epochs except for MM-IMDb which is trained for 70 epochs. All experiments are conducted on a single NVIDIA 4090 GPU with 24GB memory. 

\begin{table}[ht]
    \centering
    \vspace{-2.5mm}
    \caption{Hyperparameters for InfMasking. \textit{Masking ratio} is the ratio of masking for each masked view. 
    The V\&T CP and V\&T EE are the contact prediction and end-effector position prediction tasks on Vision\&Touch dataset respectively.
    }
    \label{tab:hyperparameters}
    \begin{tabular}{lccccc}
        \toprule
        \textit{dataset} & \textit{learning rate} & \textit{masking ratio} & \textit{number of masked views} \\
        \midrule
        \textit{Trifeature} & $3\times 10^{-4}$ & $0.7$ & $6$ \\
        \textit{MIMIC} & $3\times 10^{-4}$ & $0.8$ & $6$ \\
        \textit{MOSI} & $1\times 10^{-3}$ & $0.8$ & $5$ \\
        \textit{UR-FUNNY}(2 modalities) & $1\times 10^{-3}$ & $0.5$ & $4$ \\
        \textit{MUSTARD} & $1\times 10^{-3}$ & $0.5$ & $5$ \\
        \textit{V\&T CP}(2 modalities) & $1\times 10^{-4}$ & $0.7$ & $6$ \\
        \textit{V\&T EE} & $1\times 10^{-4}$ & $0.5$ & $4$ \\
        \textit{MM-IMDb} & $1\times 10^{-3}$ & $0.8$ & $4$ \\
        \textit{UR-FUNNY}(3 modalities) & $1\times 10^{-3}$ & $0.5$ & $4$ \\
        \textit{V\&T CP}(3 modalities) & $1\times 10^{-4}$ & $0.8$ & $5$ \\
        \bottomrule
    \end{tabular}
\end{table}

\textbf{Fusion module configuration. }
For all experiments involving InfMasking, we employ the fusion module similar to that used in CoMM~\cite{dufumier2024align}, which operates on a sequence of modality-specific embeddings and is implemented as a Transformer-based encoder layer. Specifically, the architecture consists of multi-head self-attention followed by a feed-forward network, with residual connections and layer normalization. In the bimodal setting, we use a 1-layer Transformer with 8 attention heads, while in the trimodal setting, a 2-layer Transformer with the same number of heads is adopted.
In addition, a learnable $\texttt{[CLS]}$ token is appended to the input sequence, which serves as a global representation aggregating information across modalities.

\section{Dataset Details}
\label{dataset}

\subsection{Trifeature}

The Trifeature dataset~\cite{hermann2020shapes} is designed to investigate the properties of visual neural networks and comprises three distinct features: shape, color, and texture. Each feature consists of 10 categories, resulting in 1,000 unique combinations. Of these, 800 are used for training and 200 for testing.
Each training combination is instantiated three times with random rotations applied to both shape and texture components. 
Shapes are rendered within a $128 \times 128$ bounding box, with rotation angles uniformly sampled from $[-45^{\circ}, 45^{\circ}]$, and then randomly placed within a $224 \times 224$ image canvas while ensuring full visibility. Texture and color are independently applied in the same manner. Image pairs are constructed from these instances, resulting in 10,000 training pairs and 4,096 test pairs, both sampled from the same underlying distribution.

\subsection{Multibench}

According to ~\cite{liang2021multibench}, all datasets below have been pre-processed to ensure the removal of any personally identifiable information and to safeguard user privacy (some datasets don't include any personal information, \textit{e.g.}~Vision\&Touch and MM-IMDb).

\begin{itemize}

    \item \textbf{MIMIC}~\cite{johnson2016mimic} comprises 53,423 hospital admissions from 38,597 distinct patients, spanning the years 2001 to 2012. It includes two modalities: hourly clinical measurements over a 24-hour period (represented as 12-dimensional vectors, times series modality) and static patient information such as age and gender (represented as 5-dimensional vectors, tabular modalities). The task is a binary classification problem aiming to predict whether a patient belongs to ICD-9(\textit{International Statistical Classification of Diseases and Related Health Problems}) code group 7 (460–519), which is commonly used in studies on disease classification~\cite{liang2023factorized}.
    \item \textbf{MOSI}~\cite{zadeh2016mosi} consists of 2,199 video clips collected from YouTube, designed for sentiment analysis tasks. Each sample includes video, audio signals, and corresponding text transcriptions. The original annotations range continuously from -3 to 3; following the approach in ~\cite{liang2023factorized}, these labels are binarized into positive and negative classes. The model is trained based on textual and visual modalities.
    \item \textbf{UR-FUNNY}~\cite{hasan2019ur} is constructed from 1,866 TED talk videos and comprises 16,514 samples for the task of humor detection. Each sample contains video, audio, and corresponding text transcriptions. The objective is to determine whether a given sequence is humorous, formulated as a binary classification problem. For the bi-modal setting, we use the textual and visual modalities.  
    \item \textbf{MUSTARD}~\cite{castro2019mustard} is designed for sarcasm detection and is sourced from television shows such as Friends. It contains 690 balanced utterances, each comprising video, audio, and text transcriptions, annotated as either sarcastic or non-sarcastic. In our experiments, we utilize the textual and visual modalities. 
    \item \textbf{Vision\&Touch}~\cite{lee2020making} comprises data from robotic manipulation tasks, consisting of 150 trajectories, each with 1,000 time steps. The dataset includes RGB images, depth maps, force measurements, and end-effector positions and velocities. The benchmark tasks are (1) binary classification to predict whether contact will occur in the next step and (2) regression to predict the end-effector position, evaluated using mean squared error (MSE). For the bi-modal setting, we use the visual and proprioceptive modalities.
\end{itemize}

\subsection{MM-IMDb}

Multimodal IMDb (MM-IMDb)~\cite{arevalo2017gated} is designed for movie genre prediction and comprises 25,959 films, each annotated with posters, plot summaries, genre labels, and metadata. Derived from the Movielens 20M dataset~\cite{movielensdataset}, this benchmark focuses on 23-way multi-label classification. In our experiments, we utilize the image modality (movie posters) and the text modality (plot summaries). 
While MM-IMDb is part of the Multibench benchmark~\cite{liang2021multibench}, we present it separately in our experiments, as our model is trained directly on the raw data instead of relying on the pre-processed features offered by Multibench.

\section{Broader Impact}
\label{broader_impact}

This study aims to enhance the modelling of cross-modal synergy to generate more informative multimodal representations. To ensure that \model can be deployed responsibly in real-world scenarios, we highlight several key considerations.

\textbf{Computational complexity.} \model introduces an infinite masking strategy that maximizes the mutual information between masked fused views and their unmasked counterparts, strengthening the complementarity of different modalities. However, this method inevitably increases GPU memory usage, as each additional masked view amplifies the memory footprint. We encourage future work to explore lightweight variants that can alleviate the associated computational demands. 

\textbf{Privacy and security.} As discussed in Section \ref{real_world_dataset}, the datasets used in this study span multiple domains, including healthcare, sentiment analysis and multimedia. According to~\cite{liang2021multibench}, all instances containing personal information have been rigorously anonymized and de-identified. And the Vision\&Touch and MM-IMDb datasets do not contain any personally identifiable information. 
All experiments are conducted using irreversible, pre-extracted features, except for MM-IMDb, which is processed directly from raw data; no raw or reconstructable user data is accessed, thereby minimizing privacy risks. 

\textbf{Future work.} 
Future research will focus on establishing rigorous theoretical frameworks to quantify and formally characterize the synergistic information extracted by InfMasking. Such frameworks would provide mathematical guarantees on information preservation while elucidating the fundamental limits of multimodal representation learning. Additionally, we aim to develop adaptive masking strategies that dynamically optimize masking patterns based on task requirements and modality-specific characteristics, potentially employing reinforcement learning to fine-tune these configurations. These advancements would significantly enhance our capacity to model complex synergistic relationships in multimodal data, advancing the field toward more generalizable multimodal intelligence.

\section{Additional Experiments}

\subsection{Difference with MAE}
\label{MAE_difference}

Masked Autoencoders(MAE)~\cite{he2022masked} achieve self-supervised learning through reconstruction of masked image patches, which consist of two parts. For the encoder, it encodes randomly masked  image patches into latent features. The decoder is trained to predict the masked patches using reconstruction loss, thereby enhancing semantic relationships between them in a single-modal (vision) setting. 
InfMasking adapts and extends this masking paradigm to a contrastive multimodal context. It focuses on aligning and extracting synergistic information from multimodal tokens (\textit{e.g.}, features from text, images, audio, or tabular data) through infinite masking, emphasizing cross-modal interactions like redundancy, uniqueness, and synergy. 

\begin{table}[h] 
    \centering
    \vspace{-2.5mm}
    \caption{Linear probing accuracy (\%) on three datasets from MultiBench~\cite{liang2021multibench} for MAE, CoMM, and InfMasking models.}
    \begin{tabular}{ccccc}
        \toprule
        Dataset   & MIMIC & UR-FUNNY     & MOSI    & average   \\
        \midrule
        CoMM      &$66.4_{\pm0.41}$ & $63.3_{\pm 0.51}$ & $63.7_{\pm 2.5}$ & 64.47 \\
        MAE       &$67.4_{\pm 0.3}$ & $62.5_{\pm 1.43}$ & $65.4_{\pm 1.6}$ & 65.1 \\
        InfMasking &$68.1_{\pm0.42}$ & $64.3_{\pm 0.9}$  & $69.0_{\pm 1.2}$ & 67.12 \\
        \bottomrule
        \end{tabular}
    \label{tab:restruct_results}
\end{table}

Unlike MAE, our masking approach does not mask the raw input of each modality but rather masks the features of each modality before fusion. Furthermore, we aim to maximize mutual information between masked and unmasked multimodal representations without reconstruction. It derives a lower bound approximation for the InfMasking loss assuming Gaussian distributions for masked features, making it computationally feasible for infinite views. 
This makes InfMasking a natural evolution for handling diverse modalities, addressing limitations in traditional contrastive learning (\textit{e.g.}, over-reliance on multiview redundancy) while preserving MAE's core idea of using masking to create challenging, informative views.

We further compare our InfMasking loss with MAE reconstruction loss on multiple datasets from MultiBench~\cite{liang2021multibench}. The results are illustrated in Tab.~\ref{tab:restruct_results}.
Under the same experimental conditions, we randomly mask tokens across modalities, encoder forward pass with masked iuput, and decoder-based reconstruction focused solely on masked tokens (using MSE loss averaged over masked positions). This creates a generative baseline analogous to MAE but extended to multimodal tokens. The MAE variant replaces the InfMasking loss component terms in Eq.~(\ref{eq:name-loss}) with reconstruction loss. 

\subsection{Ablation Studies of Data Augmentation on Trifeature Datasets}

The InfoMin Principle~\cite{tian2020contrastive} plays a pivotal role in self-supervised learning. It demonstrates that data augmentation is an effective strategy for adhering to this principle, as stronger data augmentations reduce mutual information to an optimal level.
In our work, we adopt the same settings for modality-specific data augmentation as outlined in CoMM~\cite{dufumier2024align}. To extend this investigation, we further explore the influence of data augmentation strategies on the bimodal Trifeature dataset~\cite{hermann2020shapes}. 

\begin{table}[h]
    \centering
    \vspace{-2.5mm}
    \caption{Impact of data augmentation on linear probing accuracy (\%) for multimodal interactions. The term "All" refers to SimCLR~\cite{chen2020simple} augmentations. InfMasking applies "All" augmentations to both modalities, consistent with CoMM.}
    \label{tab:augmentation_effect}
        \begin{tabular}{ccccccc}
            \hline
            \multicolumn{2}{c}{Augmentations} & \multirow{2}{*}{$R$} & \multirow{2}{*}{$U_1$} & \multirow{2}{*}{$U_2$} & \multirow{2}{*}{$S$} & \multirow{2}{*}{Average} \\
            \cline{1-2}
            Modality 1 & Modality 2 &  &  &  &  &  \\
            \hline
            \{All\} & $\varnothing$ & $99.78_{\pm 0.08}$ & $85.28_{\pm 2.88}$ & $49.89_{\pm 8.73}$ & $50.0_{\pm 0.0}$ & $71.24$ \\
            $\varnothing$ & \{All\} & $99.85_{\pm 0.06}$ & $49.08_{\pm 2.65}$ & $87.44_{\pm 3.59}$ & $50.0_{\pm 0.0}$ & $71.59$ \\
            \{All\} \textbackslash \{crop\} & \{All\} & $97.70_{\pm 0.84}$ & $58.07_{\pm 2.68}$ & $87.15_{\pm 3.80}$ & $50.0_{\pm 0.0}$ & $73.23$ \\
            \{All\} & \{All\} \textbackslash \{crop\} & $96.91_{\pm 1.99}$ & $85.42_{\pm 4.01}$ & $57.85_{\pm 6.63}$ & $50.0_{\pm 0.0}$ & $72.54$ \\
            \midrule
            \multicolumn{2}{c}{InfMasking} & $99.86_{\pm 0.10}$ & $90.30_{\pm 1.52}$ & $90.80_{\pm 2.88}$ & $77.02_{\pm 4.22}$ & $89.5$ \\
            \hline
        \end{tabular}
\end{table}

As shown in Tab.~\ref{tab:augmentation_effect}, omitting data augmentation leads to a significant degradation in model performance, particularly in uniqueness. Notably, cropping as a critical transformation in self-supervised learning for vision tasks~\cite{chen2020simple, he2020momentum}, is vital for learning synergistic representations in the Trifeature dataset. When modality-specific cropping augmentation is omitted, the model struggles to capture the uniqueness of the corresponding modality, resulting in an inability to effectively learn synergy.

\section{Analysis of Gaussian Approximation Assumption via Visualization}

\begin{wrapfigure}{r}{6cm}
    \centering 
    \includegraphics[width=\linewidth]{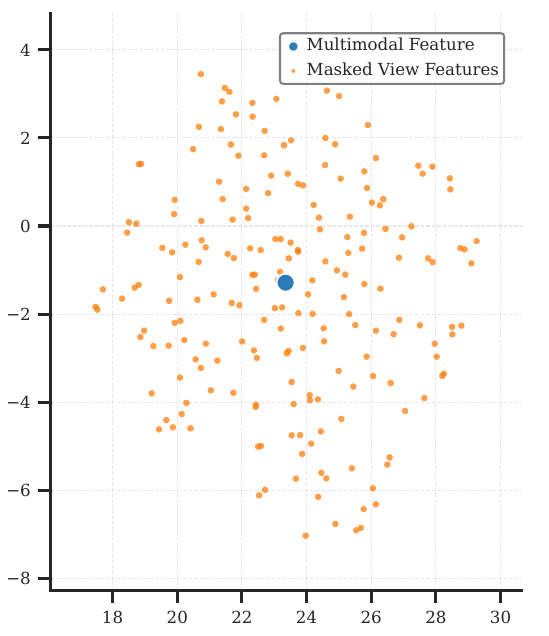}
    \caption{Visualization of the distribution of multimodal fusion embeddings and its masked counterpars.}
    \label{fig:tsne}
\end{wrapfigure}

Based on the theoretical framework of InfMasking discussed in Sec.~\ref{Contrastive_Synergistic_InfMask}, we further analyze the robustness of Gaussian approximation assumption through visualization. 

We employ dimensionality reduction to project the high-dimensional embeddings of multimodal features and their masked view features from the Trifeature dataset into a two-dimensional space, which are visualized using t-SNE.
As shown in Fig.~\ref{fig:tsne}, the masked embeddings cluster around a central point in the projected space. Notably, this central point aligns closely with the multimodal features, indicating that despite the perturbations introduced by masking, the masked embeddings inherently preserve core aspects of the synergistic semantic nature. It uncovers the stability of synergistic integration: even when parts of modalities are obscured, the fused representations converge toward a shared semantic manifold, reflecting the emergent properties that arise from modal complementarity rather than redundancy alone. 
Furthermore, the dispersion of masked embeddings similarly indicates semantic differentiation within the ambient embedding space. The observed variance highlights subtle nuances in how masking affects the differences of representations combined from masked view features, potentially corresponding to variations in semantic granularity—such as implicit biases or contextual implications that become apparent only through joint modal analysis.

\section{Pseudo-Code}
\label{pseudo_code}

Algorithm~\ref{algo} outlines the training procedure of InfMasking, formulated in the general case with $n$ modalities (\textit{e.g.}, image, text, audio, \textit{etc}).  

The key input components are as follows: $\mathcal{T}^\star$ denotes a set of label-preserving transformations used for data augmentation. The fusion transformer $g$ integrates latent features from diverse modalities. The masked view number $M'$ indicates how many masked instances are generated per modality. The random masking operator $\mathcal{M}$ stochastically obscures portions of the embedding features. And the temperature parameter $\tau$ controls the sharpness of the total loss. 

\begin{algorithm}[t]
	\caption{Multimodal contrastive interaction learning with InfMasking}
	\label{algo}
	\begin{algorithmic}[1]
		\State \textbf{Input: } Multi-modal dataset $\{X_1, X_2, ..., X_n\}$, label-preserving transformations $\mathcal{T^\star}$, set of projection transformations $\mathcal{P} = \{p_1,\ldots,p_n\}$, batch size $N$, masked view number $M'$, uni-modal encoders $(f_i)_{i\in [1..n]}$, fusion transformer $g$, random mask operator $\mathcal{M}$, temperature parameter $\tau$. 
		\For{sampled mini-batch $\{\mathbf{x}_k\}_{k\in [1..N]}=(\mathbf{x}_k^{1}, ..., \mathbf{x}_k^{n})_{k\in [1..N]}$}
            \For{$k\in [1..N]$}
            \State draw $t', t''\sim \mathcal{T^\star}$
            \State $\mathbf{x'}_i, \mathbf{x''}_i \gets t'(\mathbf{x}_i), t''(\mathbf{x}_i)$ 
            
                \For{$j \in [1..M']$}
                    \State $\mathbf{z'}_{mask}^j \gets g\left(\mathcal{M}( f_1(\mathbf{x'}_k^1) )
                    , ..., \mathcal{M}(f_n(\mathbf{x'}_k^n)) \right)$
                    \State $\mathbf{z''}_{mask}^j \gets g\left(\mathcal{M}( f_1(\mathbf{x''}_k^1) )
                    , ..., \mathcal{M}(f_n(\mathbf{x''}_k^n)) \right)$
                \EndFor
            \State $\mathbf{z'}_k \gets g(f_1(\mathbf{x'}_k^1), ..., f_n(\mathbf{x'}_k^n))$
            \State $\mathbf{z''}_k \gets g(f_1(\mathbf{x''}_k^1), ..., f_n(\mathbf{x''}_k^n))$
                \For{$i\in [1..n]$}
                        \State $\mathbf{x}_k^i \gets p_i(\mathbf{x}_k)$
                        \State $\mathbf{z}_k^i \gets g(f_i(\mathbf{x}_k^i))$
                \EndFor
		\EndFor
        \State $\mathcal{L}_\text{InfMasking} \gets -\frac{1}{M'}\sum\limits_{k=1}^{M'}\Big[ \mathbb{E}_{\substack{{z'}_{mask}^k,{z'}_{\text{pos}}\sim p({Z'}_{mask}^k,Z')}} \left[ \log \frac{\exp({{z'}_{mask}^k}^T z'_{\text{pos}}/\tau)}{\exp({{z'}_{mask}^k}^T z'_{\text{pos}}/\tau)+\sum_{z'_{\text{neg}} } \exp (z^{\prime T}z^{\prime}_{neg}/\tau)} \right]$
            \State \quad\quad\quad\quad\quad\quad\quad $+ \mathbb{E}_{\substack{{z''}_{mask}^k,z''_{\text{pos}}\sim p({Z''}_{mask}^k,Z'')}} \left[ \log \frac{\exp({{z''}_{mask}^k}^T z''_{\text{pos}}/\tau)}{\exp({{z''}_{mask}^k}^T z''_{\text{pos}}/\tau)+\sum_{z''_{\text{neg}} } \exp (z^{\prime \prime T}z^{\prime \prime}_{neg}/\tau)} \right] \Big]$
        \Statex
        
        \For{$i \in [1..n]$}
            \State $\mathcal{L}_i \gets -\Big[ \mathbb{E}_{\substack{z_i,z'_{\text{pos}}\sim p(Z_i,Z')}} \left[ \log \frac{\exp(z_i^T z'_{\text{pos}}/\tau)}{\exp(z_i^T z'_{\text{pos}}/\tau)+\sum_{z'_{\text{neg}} } \exp(z_i^T z'_{\text{neg}}/\tau)} \right]$
            \State \quad\quad $+ \mathbb{E}_{\substack{z_i,z''_{\text{pos}}\sim p(Z_i,Z'')}} \left[ \log \frac{\exp(z_i^T z''_{\text{pos}}/\tau)}{\exp(z_i^T z''_{\text{pos}}/\tau)+\sum_{z''_{\text{neg}} } \exp(z_i^T z''_{\text{neg}}/\tau)} \right] \Big]$
        \EndFor
        \Statex
        
        
        \State $\mathcal{L} \gets -\mathbb{E}_{\substack{z',z''_{\text{pos}}\sim p(Z',Z'')}} \left[ \log \frac{\exp({z'}^Tz''_{\text{pos}}/\tau)}{\exp({z'}^Tz''_{\text{pos}}/\tau)+\sum_{z''_{\text{neg}} } \exp({z'}^T,z''_{\text{neg}}/\tau)} \right]$
        \Statex
        \Statex
        \State $\mathcal{L}_\text{Total~loss} \gets \mathcal{L} + \sum_{i=1}^{n}\mathcal{L}_i + \mathcal{L}_\text{InfMasking}$
        \State update $(f_i)_{i\in [1..n]}, \mathcal{M}, g$ to minimize $\mathcal{L}_{\text{Total~loss}}$
		
		\EndFor
        \State \textbf{return} $(f_i)_{i\in [1..n]}, g$
	\end{algorithmic}
\end{algorithm}

\section{Proof}\label{aprooof}
\begin{proof}[\cref{lemma: ifif}]
\begin{align}
   & \mathbb{E}_{\text{mask}}[\hat{I}_{\text{NCE}}(Z^{\prime}_{\text{mask}}, Z^{\prime})] \\ &=\mathbb{E}_{\text{mask}}[\mathbb{E}_{z'\sim p(Z')}\left[ \log \frac{\exp (z'^Tz^{\prime}_{{\text{mask}}}/\tau)}{\exp (z'^Tz^{\prime}_{{\text{mask}}}/\tau)+\sum_{z'_{\text{neg}} } \exp (z^{\prime T}z^{\prime}_{neg}/\tau)} \right] ] \\ &=\mathbb{E}_{z'\sim p(Z')}[\mathbb{E}_{\text{mask}}\left[ \log \frac{\exp (z'^Tz^{\prime}_{{\text{mask}}}/\tau)}{\exp (z'^Tz^{\prime}_{{\text{mask}}}/\tau)+\sum_{z'_{\text{neg}} }\exp (z^{\prime T}z^{\prime}_{neg}/\tau)} \right] ]\\ &
   =\mathbb{E}_{z'\sim p(Z')}[\mathbb{E}_{\text{mask}}\left[ (z'^Tz^{\prime}_{{\text{mask}}}/\tau)-\log[ \exp (z'^Tz^{\prime}_{{\text{mask}}}/\tau)+\sum_{z'_{\text{neg}} } \exp (z^{\prime T}z^{\prime}_{neg}/\tau)] \right] ]
   \\ &\geq\mathbb{E}_{z'\sim p(Z')}\left[ z^{\prime T}\mathbb{E}_{\text{mask}}[z^{\prime}_{{\text{mask}}}]/\tau-\log \mathbb{E}_{\text{mask}}[ \exp (z'^Tz^{\prime}_{{\text{mask}}}/\tau)+\sum_{z'_{\text{neg}} } \exp (z^{\prime T}z^{\prime}_{neg}/\tau)] \right ]\label{eq:inequality4}
\end{align}
The inequality Eq.(\ref{eq:inequality4}) merges from the application of Jensen inequality on concave functions i.e., $\mathbb{E}_x \log (X) \leq \log \mathbb{E}_x[X]$. 
$z^{\prime}_{{\text{mask}}}$ denotes the integrated representation derived from the fusion of all masked features across diverse modalities via the Transformer architecture. 
 \begin{lemma}\label{lemma: MGF}
    Consider a random variable $\boldsymbol{x}$ that follows a multivariate Gaussian distribution, denoted as $\boldsymbol{x} \sim \mathcal{N}(\boldsymbol{\mu}, \boldsymbol{\Sigma})$, where $\boldsymbol{\mu} \in \mathbb{R}^n$ represents the mean vector and $\boldsymbol{\Sigma} \in \mathbb{R}^{n \times n}$ is the covariance matrix 
 The moment generating function (MGF) of this random variable is given by the following expression:
     \begin{equation}\label{eq: MGF}
        \mathbb{E}_{\boldsymbol{x}}\left[e^{\boldsymbol{a}^T \boldsymbol{x}}\right]=e^{\boldsymbol{a}^T \boldsymbol{\mu}+\frac{1}{2} \boldsymbol{a}^T \boldsymbol{\Sigma} \boldsymbol{a}},
    \end{equation}
    where $\boldsymbol{a} \in \mathbb{R}^n$ is an arbitrary constant vector. 
\end{lemma}
$z'^Tz^{\prime}_{{\text{mask}}}/\tau$
According to Lemma \ref{lemma: MGF}, we can derive the MGF of the inequality Eq.(\ref{eq:inequality2}) as follows:
\begin{align}
    & \mathbb{E}_{\text{mask}}[\hat{I}_{\text{NCE}}(Z^{\prime}_{\text{mask}}, Z^{\prime})] \\ &\geq\mathbb{E}_{z'\sim p(Z')}\left[ z'^T\boldsymbol{\mu}_{z_{\text{mask}}^{\prime}} /\tau-\log [ \exp (z'^T\boldsymbol{\mu}_{z_{\text{mask}}^{\prime}} /\tau+\frac{1}{2\tau^2}z'^T\boldsymbol{\Sigma}_{z_{\text{mask}}^{\prime}}z)+\sum_{z'_{\text{neg}} } \exp (z^{\prime T}z^{\prime}_{neg}/\tau)] \right ]\label{eq:inequality}
 \end{align} 
 \end{proof}




\end{document}